\documentclass[letterpaper, 10 pt, conference, twoside]{IEEEtran}  

\IEEEoverridecommandlockouts                                                              
\usepackage[english]{babel}
\usepackage{times} 
\usepackage{cite}
\usepackage[font=footnotesize]{subcaption}
\usepackage{calrsfs}
\DeclareMathAlphabet{\pazocal}{OMS}{zplm}{m}{n}
\usepackage{balance} 

\usepackage[nolist]{acronym}

\usepackage{cellspace}
\newcolumntype{D}{>{\hfill}N{3}{2}<{\hfill}}

\usepackage{xcolor}
\usepackage[%
	colorlinks = false,
	linkcolor = blue,
	urlcolor  = blue,
	citecolor = blue,
	anchorcolor = blue]{hyperref}

\def\xx{\mathsf{x}}
\def\yy{\mathsf{y}}
\def\zz{\mathsf{z}}

\usepackage{pifont}

\usepackage{graphicx} 
\usepackage{epsfig} 
\usepackage{pgfplots}
\usepackage{caption}
\usepackage{subcaption}
\usepackage{booktabs}

\pgfplotsset{compat=1.15}
\graphicspath{{./figure/}}

\usepackage[export]{adjustbox}

\makeatletter
\let\MYcaption\@makecaption
\makeatother

\makeatletter
\let\@makecaption\MYcaption
\makeatother

\usepackage{mathptmx} 
\usepackage{amsmath} 
\usepackage{amssymb}  
\usepackage{nicefrac}
\usepackage{mathtools}
\usepackage{optidef}
\usepackage{leftidx}
\usepackage{bm} 

\DeclareMathAlphabet{\pazocal}{OMS}{zplm}{m}{n}

\def\x#1{\texttt{\expandafter\string\csname#1\endcsname}&\expandafter$\csname#1\endcsname$}

\makeatletter%
\AfterPreamble{%
	\usepackage{hyperref}%
	\let\oldhypertarget\hypertarget%
	\renewcommand{\hypertarget}[2]{%
		\oldhypertarget{#1}{#2}%
		\protected@write\@mainaux{}{%
			\string\expandafter\string\gdef%
			\string\csname\string\detokenize{#1}\string\endcsname{#2}%
		}%
	}%
	\newcommand{\myhyperlink}[1]{%
		\hyperlink{#1}{\csname #1\endcsname}%
	}%
}
\makeatother%

\newcounter{Definition}
\newcounter{Theorem}
\makeatletter
\providecommand{\bigsqcap}{%
	\mathop{%
		\mathpalette\@updown\bigsqcup
	}%
}
\newcommand*{\@updown}[2]{%
	\rotatebox[origin=c]{180}{$\m@th#1#2$}%
}
\makeatother

\DeclareMathOperator*{\minimize}{minimize}
\DeclareMathOperator*{\maximize}{maximize}

\usepackage{float}
\usepackage{tikz}
\pgfdeclarelayer{foreground}
\pgfsetlayers{background, main, foreground}
\usetikzlibrary{quotes, angles, backgrounds, arrows, automata, shapes, positioning, calc, through, spy, decorations.pathreplacing, decorations.markings, arrows.meta, 3d, perspective, petri, intersections, fillbetween, patterns, pgfplots.fillbetween}

\tikzset{
    imglabel/.style={
      rectangle,
      inner sep=2pt,
      text=black,
      minimum height=1em,
      text centered,
      fill=white,
      fill opacity=1.0,
      text opacity=1,
      anchor=south west,
    },
  }

\tikzset{
	state/.style={
		rectangle,
		draw=black, very thick,
		minimum height=1.0em,
		text centered,
	},
}

\definecolor{redRSE}{RGB}{192,0,0}
\definecolor{darkgreen}{rgb}{0.272, 0.50, 0.376}
\definecolor{lightgreen}{rgb}{0.585, 0.82, 0.647}

\def\tilt{55}      
\def\lobescale{1.5}

\definecolor{recvcol}{RGB}{220,0,70}
\definecolor{zcol}{RGB}{10,120,10}
\definecolor{axisgray}{gray}{0.55}

\usepackage{siunitx}
\usepackage{todonotes}

\usepackage{algorithm}
\usepackage[noend]{algpseudocode}

\makeatletter
\def\BState{\State\hskip-\ALG@thistlm}
\makeatother

\title{ \LARGE \bf
        Communications-Aware NMPC for Multi-Rotor Aerial Relay Networks Under Jamming Interference
}

\author{Giuseppe Silano$^{1,2}$, Daniel Bonilla Licea$^{3,1}$, Davide Liuzza$^{4}$, Antonio Franchi$^{5}$, and Martin Saska$^{1}$
    \thanks{This work was partially supported by the GA\v{C}R project no. 26-22419S, the CTU grant no. SGS26/077/OHK3/1T/13, the EU project ``Robotics and Advanced Industrial Production'' (reg.\ no.\ CZ.02.01.01/00/22008/0004590), the EU project ``AUTOASSESS'' (no.\ 101120732), and by the research fund for the Italian Electrical System (decree n.\ 388 of  November 6th, 2024).} 
    \thanks{$^1$Department of Power Generation Technologies and Materials, Ricerca sul Sistema Energetico S.p.A., 20134 Milan, Italy.}
    \thanks{$^2$Department of Cybernetics, Czech Technical University in Prague, 12135 Prague, Czech Republic (e-mails: {\tt\small \{giuseppe.silano, martin.saska\}@fel.cvut.cz}).}
    \thanks{$^3$College of Computing, Mohammed VI Polytechnic University, 11103, Morocco (e-mail: {\tt\small daniel.bonilla@um6p.ma}).}
    \thanks{$^4$Department of Engineering, University of Sannio in Benevento, 82100 Benevento, Italy, 
    (e-mail: {\tt\small dliuzza@unisannio.it}).}
    \thanks{$^5$Robotics and Mechatronics Department, University of Twente, 7500 AE Enschede, The Netherlands, and also with the Department of Computer, Control and Management Engineering, Sapienza University of Rome, 00185 Rome, Italy (e-mail: {\tt\small schol@r-franchi.eu}).}
}

\begin{document}

\maketitle
\thispagestyle{empty}
\pagestyle{empty}

	
\begin{acronym}
    \acro{AR}[AR]{Aerial Relay}
    \acro{AHP}[AHP]{Analytic Hierarchy Process}
    \acro{AP}[AP]{Access Point}
    \acro{AoA}[AoA]{Angle of Arrival}
    \acro{AoD}[AoD]{Angle of Departure}
    \acro{AWGN}[AWGN]{Additive White Gaussian Noise}
    \acro{BS}[BS]{Base Station}
    \acro{CCW}[CCW]{Counter-ClockWise}
    \acro{CSI}[CSI]{Channel State Information}
    \acro{CW}[CW]{ClockWise}
    \acro{CoM}[CoM]{Center of Mass}
    \acro{DRL}[DRL]{Deep Reinforcement Learning}
    \acro{FEC}[FEC]{Forward Error Correction}
    \acro{FFD}[FFD]{Frequency-Division Duplexing}
    \acro{FoV}[FoV]{Field of View}
    \acro{GNSS}[GNSS]{Global navigation satellite system}
    \acro{GTMR}[GTMR]{Generically-Tilted Multi-Rotor}
    \acro{IoT}[IoT]{Internet of Things}
    \acro{IRS}[IRS]{Intelligent Reflecting Surface}
    \acro{LoS}[LoS]{Line of Sight}
    \acro{MPC}[MPC]{Model Predictive Control}
    \acro{MRAV}[MRAV]{Multi-Rotor Aerial Vehicle}
    \acro{NMPC}[NMPC]{Nonlinear Model Predictive Control}
    \acro{NLP}[NLP]{Nonlinear Programming Problem}
    \acro{OCC}[OCC]{Optical Camera Communication}
    \acro{OFDMA}[OFDMA]{Orthogonal Frequency Division Multiple Access}
    \acro{OS}[OS]{Operating System}
    \acro{OWC}[OWC]{Optical Wireless Communication}
    \acro{QP}[QP]{Quadratic Programming}
    \acro{RF}[RF]{Radio Frequency}
    \acro{RL}[RL]{Reinforcement Learning}
    \acro{RSS}[RSS]{Received Signal Strength}
    \acro{SNR}[SNR]{Signal-to-Noise Ratio}
    \acro{SINR}[SINR]{Signal-to-Interference-plus-Noise Ratio}
    \acro{SQP}[SQP]{Sequential Quadratic Programming}
    \acro{TDD}[TDD]{Time Division Duplex}
    \acro{TDoA}[TDoA]{Time Difference of Arrival}
    \acro{UAV}[UAV]{Unmanned Aerial Vehicle}
    \acro{UVDAR}[UVDAR]{UltraViolet Direction and Ranging}
    \acro{UV}[UV]{Unmanned Vehicle}
    \acro{wrt}[w.r.t]{with respect to}     
\end{acronym}
	


\begin{abstract}

    \acp{MRAV} are increasingly used in communication-dependent missions where connectivity loss directly compromises task execution. Existing anti-jamming strategies often decouple motion from communication, overlooking that link quality depends on vehicle attitude and antenna orientation. In coplanar platforms, ``tilt-to-translate'' maneuvers can inadvertently align antenna nulls with communication partners, causing severe degradation under interference.    
    This paper presents a modular communications-aware control approach combining a high-level $\max$-$\min$ trajectory generator with an actuator-level Nonlinear Model Predictive Controller (\acs{NMPC}). The trajectory layer optimizes the \emph{weakest link}---the link with the lower instantaneous \acl{SINR}, which bottlenecks the harmonic-mean end-to-end capacity---while the \acs{NMPC} enforces vehicle dynamics, actuator limits, and antenna-alignment constraints. Antenna directionality is handled geometrically, avoiding explicit radiation-pattern parametrization. 
    The method is evaluated in a relay scenario with an active jammer and compared across coplanar and tilted-propeller architectures. Results show a near two-order-of-magnitude increase in minimum end-to-end capacity, markedly reducing outage events, with moderate average-capacity gains. Tilted platforms preserve feasibility and link quality, whereas coplanar vehicles show recurrent degradation. These findings indicate that full actuation is a key enabler of reliable communications-aware operation under adversarial directional constraints.

\end{abstract}






\section{Introduction}
\label{sec:introduction}

Recent advances in sensing, communications, computation, and actuation have enabled autonomous robot teams with advanced perception, interaction, and communication capabilities \cite{Muralidharan2021ARCRAS, Zeng2019ProceedingsIEEE, Bonilla2024ProcIEEE, Hurst2024}, driving deployment in surveillance \cite{Mozaffari2019CST}, cellular network extension \cite{ZengWC2019}, infrastructure monitoring \cite{Shakeri2019IEEECST}, and data collection~\cite{Stachura2017JFR}.

\acf{RF} communication is central to these networks, but maintaining reliable connectivity under mobility is challenging \cite{Muralidharan2021ARCRAS, Zeng2019ProceedingsIEEE, Bonilla2024ProcIEEE}, and these issues become more severe in adversarial settings where \ac{RF} links are exposed to eavesdropping and jamming \cite{ZhongIEEECommLetters2019, JiIEEETVT2022, SilanoSMC2025}. 
\acp{UAV} are particularly exposed because their altitude and geometry often create \ac{LoS} links with ground-based jammers \cite{BonillaICASSP2024}, leaving limited shielding or rerouting options. Classical countermeasures, including frequency hopping, spectrum spreading, and transmit power control \cite{VADLAMANI2016IJPE}, provide only partial mitigation in contested environments.

\acfp{MRAV} are particularly versatile rotary-wing platforms, capable of controlling position and orientation while benefiting from mechanical simplicity and redundant actuation \cite{RiberioWONS2025}. Coplanar configurations (e.g., quadrotors) offer a balance of simplicity and control authority, whereas tilted-propeller designs provide additional actuation freedom at the expense of higher energy consumption \cite{Ryll2019IJRR, BonillaIEEEComMag2025}.

While \acp{MRAV} are inherently vulnerable to \ac{RF} jamming, their agility makes them promising tools for active mitigation \cite{ZhongIEEECommLetters2019, JiIEEETVT2022}: they can adapt position and orientation in real time to improve link quality and preserve connectivity, enabling communications-aware behaviors beyond conventional strategies \cite{BonillaIEEEComMag2025}. Despite extensive work on trajectory design and anti-jamming techniques, most approaches still model \acp{MRAV} as kinematic points with isotropic or weakly directional radios \cite{Bonilla2024ProcIEEE, ChenVTCF2018}, overlooking a fundamental physical coupling: lateral motion requires body tilting, which directly reorients directional antennas---so geometrically optimal maneuvers can drive antenna nulls toward communication nodes, yielding trajectories that are dynamically feasible but communication-fragile on real platforms.

The field of \textit{communications-aware robotics} \cite{Muralidharan2021ARCRAS, Zeng2019ProceedingsIEEE, Bonilla2024ProcIEEE} addresses these challenges by jointly embedding communication objectives---mobility, connectivity, sensing, and energy---into motion planning and control.

This paper introduces a modular communications-aware control architecture for \acp{MRAV}, composed of a \textit{trajectory generator} and a \textit{Nonlinear Model Predictive Controller} (\acs{NMPC}). The trajectory generator solves a $\max$--$\min$ optimization problem to strengthen the weakest communication link under jamming;  
the \acs{NMPC} then tracks these references while enforcing actuator constraints, communication alignment, and dynamic feasibility. A key feature is that antenna directionality is incorporated via geometric alignment constraints rather than explicit antenna radiation models.
The architecture applies to both coplanar and tilted-propeller \acp{MRAV} \cite{Ryll2019IJRR}.
As a case study, it is instantiated in a relay scenario in which a \ac{MRAV} acts as an aerial relay between a ground base station and a mobile peer under active jamming (see Figure \ref{fig:relayScenario}), used to evaluate end-to-end communication performance and operational reliability under adversarial conditions.



\section{Related work}
\label{sec:relatedWork}

Research on \ac{UAV}-based wireless networks in adversarial environments has focused mainly on \textit{trajectory optimization} and \textit{anti-jamming techniques} \cite{LohanIEEEOJCS2024}, but these are usually treated separately, overlooking that \acp{MRAV} can jointly exploit mobility, attitude, and communication as a coupled resource.



\subsection{Trajectory optimization}
\label{sec:trajectoryOptimization}

Trajectory optimization adapts \ac{UAV} flight paths to improve link quality, reduce outages, and avoid interference, spanning graph-theoretic planners, convex optimization, and \ac{RL} \cite{Zhang2019IEEETC, Bulut2018ICCWorkshop, Khamidehi2020ICCWorkshop, Zeng2021IEEETWC, ChallitaIEEETWC2019}. Specific examples include connectivity-maximizing planners, dynamic programming, and \ac{RL}-based policies for reducing out-of-coverage time \cite{Zhang2019IEEETC, Bulut2018ICCWorkshop, Khamidehi2020ICCWorkshop}; \ac{DRL} has also been used to reduce mission time and outages \cite{Zeng2021IEEETWC}, and interference-aware planning to balance energy, latency, and interference \cite{ChallitaIEEETWC2019}.

Joint trajectory–power optimization has been formulated as Stackelberg games \cite{XuICC2019}, and iterative methods such as block coordinate descent and successive convex approximation have been used to construct jamming-resilient paths \cite{Bulut2018ICCWorkshop, Khamidehi2020ICCWorkshop}. Extensions include multiple-jammer scenarios \cite{Bulut2018ICCWorkshop}, 3D planning with turning/climbing constraints \cite{Khamidehi2020ICCWorkshop}, and \ac{UAV}-to-\ac{UAV} links under adversarial interference \cite{DuoIEEETVT2020, YangIEEEWCSP}.

Despite this progress, most trajectory-optimization frameworks embed communication metrics at the planning problem but neglect their interaction with vehicle-level dynamics and control constraints \cite{Bonilla2024ProcIEEE, ChenVTCF2018}, motivating integrated methods that jointly handle nonlinear dynamics, actuator limits, and communication objectives to ensure reliable execution in adversarial environments. 



\subsection{Anti-jamming techniques}
\label{sec:uavAndJammers}

Anti-jamming strategies can be broadly grouped into retreat strategies and physical-layer countermeasures.

\textit{Retreat strategies} exploit \acp{UAV} mobility to evade jammed regions with minimal hardware complexity; classic examples include channel surfing and spatial retreating \cite{Xu2004ACMWorkshop, LiIEEEAccess2019}. Game-theoretic formulations have been used to compute evasive trajectories \cite{Bhattacharya2010ACC}, while geometric tools such as Voronoi diagrams \cite{Judd2001AIAA} and power-based threat maps \cite{Duan2014ICSP} provide lightweight heuristics. However, these approaches typically rely on kinematic models and neglect actuator limits and aerodynamic couplings, leading to trajectories that can be dynamically infeasible or too slow for rapidly changing conditions.

\textit{Physical-layer countermeasures} attempt to withstand jamming rather than evade it. Typical techniques include transmit power control to boost \ac{SINR} \cite{Shichao2017GLOBECOM} and error-correcting codes to mitigate packet loss \cite{WenyuanIEEENetwork2006}; more recent work applies \ac{DRL} for real-time adaptive power allocation \cite{LuIEEEWC2020}. Although these approaches improve resilience, they often assume constant-power, always-on jammers---limiting applicability against adaptive adversaries---and introduce energy or throughput trade-offs that are critical for resource-constrained \ac{UAV} platforms.

Despite their differences, both classes largely treat motion and communication as independent: 
retreat methods act on geometry but neglect vehicle dynamics, while physical-layer methods operate at the signal level and overlook antenna orientation and motion-induced effects \cite{BonillaEUSIPCO21}, so neither class fully leverages actuation geometry and attitude as communication-control variables.

Overall, existing methods either simplify vehicle dynamics or omit antenna directionality, preventing communication objectives from being enforced at the control level and motivating integrated frameworks that embed vehicle dynamics, actuation constraints, and communication objectives into a unified control formulation. 



\subsection{Positioning against the literature}
\label{sec:positioningLiterature}

Table~\ref{tab:related-work} substantiates this gap across the works discussed in Sections~\ref{sec:trajectoryOptimization} and~\ref{sec:uavAndJammers}. ``Veh.'' is ``Kin'' for a kinematic or point-mass model and ``Dyn'' for a full nonlinear dynamic model; ``Act.'' is ``Iso'' for isotropic, ``Dir'' for an explicit directional pattern, and ``Geom'' for the geometric-alignment treatment adopted here; ``Jam.'' is ``Iso'' for an isotropic jammer model or a dash if none is modeled; ``RH'' indicates a receding-horizon implementation; ``E2E'' indicates an end-to-end relay setting; ``Coupl.'' indicates explicit coupling of attitude and actuation to link quality, with ``static'' and ``partial'' denoting coupling only at a fixed pose or without a receding-horizon controller, respectively. 

\begin{table}[t]
    \caption{Positioning of this work against representative \ac{UAV} trajectory-optimization and anti-jamming literature.}
    \label{tab:related-work}
    \vspace{-1.0em}
    \centering
    \includegraphics[width=\columnwidth]{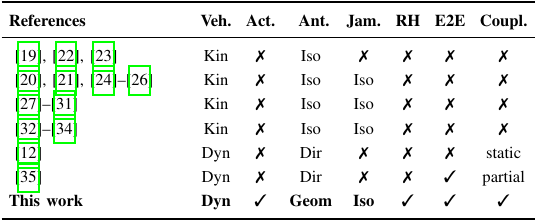}
    \vspace{-1.35em}
\end{table}

As Table~\ref{tab:related-work} shows, to the best of the authors' knowledge no prior work jointly enforces actuator feasibility and directional-alignment-aware communication constraints within a single receding-horizon controller for relay \acp{MRAV} under jamming.

Two further observations follow. First, the dominant prior methods model the platform kinematically and the antenna as isotropic, decoupling motion from link quality. Second, the works closest in spirit~\cite{BonillaICASSP2024, BonillaEUSIPCO21} account for antenna directionality but operate either statically~\cite{BonillaICASSP2024} or as one-shot planners without actuator-rate constraints, a jammer model, or receding-horizon replanning~\cite{BonillaEUSIPCO21}. Crucially, embedding the antenna-alignment constraint inside a receding-horizon dynamic controller is non-trivial: it requires a simplified yet representative two-stage problem in which communication constraints are replaced by their geometric counterparts to enable real-time computation, as detailed later.



\section{Contributions}
\label{sec:contributions}

This paper proposes a modular communications-aware control approach for \acp{MRAV} combining two tightly integrated components---a trajectory generator and a communications-aware \acs{NMPC}---to maintain reliable connectivity under jamming. 

The \textit{trajectory generator} computes high-level motion references by solving a $\max$-$\min$ optimization problem that strengthens the weakest communication link under jamming. 
To address non-convexity, the method employs a $h$-norm approximation, second-order Taylor expansions, and a quadratic reformulation, enabling efficient gradient-based optimization. Bandwidth allocation is selected to balance the link capacities and maximize end-to-end throughput.

The \textit{communications-aware \acs{NMPC}} serves as the inner-loop controller, generating rotor-speed commands while enforcing actuator constraints and dynamic feasibility; unlike conventional approaches, it jointly optimizes motion and communication objectives by embedding antenna alignment directly into the cost and constraints.  
A distinctive feature is its geometric treatment of antenna directionality, which avoids explicit radiation-pattern models and preserves alignment during agile maneuvers.  

The framework is compatible with a wide range of multi-rotor designs, from coplanar to tilted-propeller platforms \cite{Ryll2019IJRR}, and supports antennas with doughnut-shaped radiation patterns similar to that in Figure \ref{fig:doughnutShape}, enabling deployment across platforms with varying levels of agility and actuation authority. 

\begin{figure}[tb]
    \centering
    \scalebox{0.95}{
    \begin{tikzpicture}
    \tikzset{->-/.style={decoration={markings, mark=at position #1 with {\arrow{>}}}, postaction={decorate}}
    }

    \node (p_MRAV1) at (-0.40,0.25) {\scriptsize $\mathbf{p}_1$};
    \node (p_MRAV2) at (4.55,0.25) {\scriptsize $\mathbf{p}_2$};
    \node (p_BS) at (-0.65,-3.0) {\scriptsize $\mathbf{p}_0$};
    \node (p_J) at (2.65,-2.35) {\scriptsize $\mathbf{p}_J$};

    \node (MRAV1)  at (0,0) [text centered, rotate=20, overlay]{\includegraphics[scale=0.10]{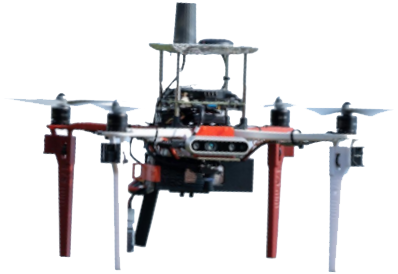}};
    \node at (-0.90,0.75) [text centered, font=\scriptsize]{MRAV-1};
    
    \node (MRAV2) at (4,0) [text centered, rotate=-20]{\includegraphics[scale=0.10]{figure/single_uav_follower_v2.png}}; 
    \node at (3.70,0.95) [text centered, font=\scriptsize]{MRAV-2};    
    
    \draw[blue, dash dot, latex-] (0.80, 0) arc (-5:120:0.65cm) node at (0.05, 0.225) 
    [above]{\scriptsize $\vartheta^A_{2,1}$};
    \draw[blue, dash dot, -latex] (-0.25, 0.725) arc (120:-40:0.90cm) node at (1.175, 0.25) 
    [above]{\scriptsize $\vartheta^A_J$};
    
    \draw[blue, dash dot, -latex] (-0.3, 0.625) arc (118:230:0.90cm) node at (-1, -0.75) [above]{\scriptsize $\vartheta^D_{1,0}$};
    
    \draw[blue, dash dot, latex-] (3.15, 0) arc (180:60:0.7cm) node at (3.0, 0.25) 
    [above]{\scriptsize $\vartheta^D_{2,1}$};
    
    \node (circle) at (MRAV1) [circle, draw, scale=0.2, fill=black] {};
    \node (circle) at (MRAV2) [circle, draw, scale=0.2, fill=black] {};
    \node (BS) at (-1.2,-2.55) [text centered, overlay]{\includegraphics[scale=0.15]{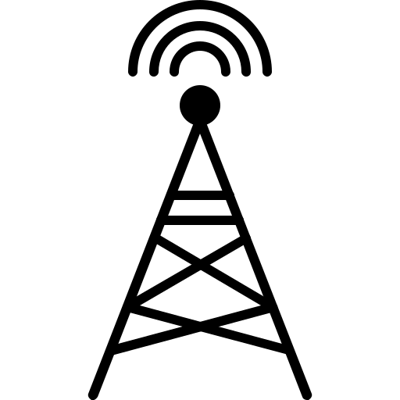}};
    \draw (-1.1,-2.25) node[right]{\scriptsize $\text{BS}$};
    \draw[-latex, solid, black] ($(MRAV2) - (0.65,0)$) -- ($(MRAV1) + (0.6,0)$);
    \draw[solid, black, -latex] ($(BS) + (0,0.8)$) -- ($(BS) + (0,1.35)$) node[left, font=\scriptsize]{$\zz_{BS}$};
    \draw[blue, dash dot, latex-, rotate=-15] ($(BS) + (0.05,1.05)$) arc (75:120:0.45cm) node at ($(BS) + (-0.05,1.15)$) [above]{\scriptsize $\vartheta^A_{1,0}$};

    \node (GroundJammer) at (3.0,-2.25) [text centered, overlay]{\includegraphics[scale=0.10]{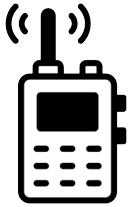}};
    \node (GroundJammer-Text) at (3.75,-2.35) [text centered, text width=5em, font=\scriptsize]{Ground\\Jammer};
    \draw[-latex] (GroundJammer) -- ($(GroundJammer) + (0,1.0)$) node[left, font=\scriptsize]{$\zz_J$};
    
    \draw[solid, black, -latex] ($(MRAV1) - (0.2,0.5)$) -- (-1,-1.75); 
    
    \draw[-latex] (MRAV2) node at (3.90,-0.15) [below, font=\scriptsize]{$O_{B_2}$} -- ($ (MRAV2) + (0.342020143, 0.939692621)$) node[right, font=\scriptsize]{$\zz_{B_2}$};
    \draw[-latex] ($ (MRAV2) + (0.342020143,-0.125)$) -- (MRAV2);
    \node at ($(MRAV2) + (0.75,-0.7)$) [above, font=\scriptsize]{$\xx_{B_2}$};
    \draw[-latex] (MRAV1) -- ($ (MRAV1) + (-0.342020143, 0.939692621)$) node[right, font=\scriptsize]{$\zz_{B_1}$};
    \draw[-latex] ($ (MRAV1) + (0.342020143,0.125)$) -- (MRAV1);
    \node at ($(MRAV1) + (0.45,0.2)$) [above, font=\scriptsize]{$\xx_{B_1}$};
    \node at (0.25,-0.4) [text centered, font=\scriptsize]{$O_{B_1}$};
    
    \draw[-latex] (-3.25,-3.25) node[below]{\scriptsize $O_W$} -- (-3.25,-2.25) 
    node[right]{\scriptsize $\zz_W$};
    \draw[-latex] (-3.25,-3.25) -- (-2.25,-3.25) node[below]{\scriptsize $\xx_W$};
    \node (circle) at (-3.25,-3.25) [circle, draw, scale=0.2, fill=black] {};

    \draw[redRSE, loosely dashdotdotted, -latex] (GroundJammer) -- (MRAV1);
    \draw[redRSE, loosely dashdotdotted, -latex] (GroundJammer) -- (BS);

    \draw[-latex, solid, black] (-2.5,1.35) -- (-1.5,1.35) node [right, font=\scriptsize, text width=5em]{Legitimate Links};
    \draw[-latex, loosely dashdotdotted, redRSE] (1.0,1.35) -- (2.0,1.35) node [right, font=\scriptsize, text width=5em]{Jamming Links};
    
    \end{tikzpicture}
    }
    \vspace*{-0.80em}
    \caption{Schematic of \ac{MRAV}-1 serving as a communication relay between \ac{MRAV}-2 and a ground \ac{BS} under jamming interference, with body frames $\pazocal{F}_{B_1}$, $\pazocal{F}_{B_2}$ and world frame $\pazocal{F}_W$.}
    \label{fig:relayScenario}
    \vspace*{-1.35em}
\end{figure}

Specifically, this paper considers a scenario in which relay \ac{MRAV}-1 maintains connectivity between a ground \ac{BS} and a mobile peer (\ac{MRAV}-2) under persistent, time-varying jamming (see Figure~\ref{fig:relayScenario}), used to assess end-to-end communication performance and operational reliability under adversarial conditions. 

The main contributions of the paper are:

\begin{itemize}
    \item A $\max$--$\min$ trajectory generator reduced to a quadratic, real-time-tractable surrogate via a conservative channel relaxation and a second-order Taylor approximation (see Section~\ref{sec:trajectoryGenerator}), whose output feeds the \acs{NMPC} as a time-varying reference rather than an open-loop plan.
    \item A \emph{mutual} directional-alignment constraint \eqref{eq:directionalCosine}---jointly coupling both relay links through a product-form inequality with slack-based feasibility recovery---embedded directly inside the receding-horizon \acs{NMPC} (see Section~\ref{sec:communicationAwareNMPC}), rather than enforced as two independent per-link bounds or at the planning layer alone.
    \item A unified evaluation, on a common \ac{GTMR} dynamic model, of coplanar and tilted-propeller actuation (see Section~\ref{sec:simulationResults}), isolating actuation geometry---as opposed to algorithmic choices---as the dominant factor governing communication reliability under jamming.
\end{itemize}

To the best of our knowledge, this is the first receding-horizon controller that jointly enforces actuator feasibility and a mutual directional-alignment constraint for relay \acp{MRAV} under jamming, as summarized in Table~\ref{tab:related-work}.

\section{Problem Description}
\label{sec:problemDescription}

We consider a communication relay scenario in which \ac{MRAV}-1 acts as an airborne relay between a mobile \ac{MRAV}-2 and a ground \ac{BS} under jamming interference (see Figure \ref{fig:relayScenario}), with the objective of computing dynamically feasible trajectories that preserve end-to-end connectivity under adversarial conditions.

The system integrates two complementary communication technologies: a \ac{RF} link for high-throughput data transmission and an \ac{OCC} channel for inter-agent coordination. The \ac{RF} link enables \ac{MRAV}-2 to transmit data to \ac{MRAV}-1, which relays it to the \ac{BS} \cite{ChowdhuryIEEECST2020}, a configuration required when direct \ac{MRAV}-2-to-\ac{BS} communication is infeasible due to range limits or \ac{LoS} obstructions. Although the \ac{RF} path provides mission-level throughput, it remains vulnerable to jamming, especially from ground-based adversaries with unobstructed \ac{LoS}. 

\begin{figure}
    \begin{tikzpicture}
    
        \newcommand{\DipoleSurf}[1]{%
          \addplot3 [
            surf, shader=flat, draw=black!30, line width=0.05pt, opacity=0.98,
            domain=0:180, y domain=0:360,
            samples=28, samples y=36,
            point meta={100*(sin(x))},         
            ] (
            {#1*(sin(x))^3*cos(y)},
            {#1*(sin(x))^3*sin(y)},
            {#1*(sin(x))^2*cos(x)}
          );
        }
        
        \begin{axis}[
          colormap/viridis,
          at={(0.0cm,0)}, anchor=center,
          width=5.8cm, height=4.8cm,
          axis lines=none, view={35}{25}, z buffer=sort,
        ]
        \DipoleSurf{\lobescale}

        
        \addplot3[thick, color=black] coordinates {(0,0,-1.4) (0,0,-1.0)};
        \addplot3[dashed, color=black] coordinates {(0,0,-0.8) (0,0,0.6)};
        \addplot3[-latex, solid, color=black] coordinates {(0,0,0.6) (0,0,1.2)};
        \node[anchor=west, text=black, font=\scriptsize] at (axis cs:0,0,1.2) {$\zz_{B_2}$};

        \node[anchor=west, text=black, font=\scriptsize] at (axis cs:-0.30,0,-2.2) {(a)};
        
        \addplot3[-latex, color=redRSE] coordinates {(0,0,0) (\lobescale*1.02,0,0)};
        \node[redRSE, anchor=north, font=\scriptsize] at (axis cs:\lobescale*1.05,0,0) {\ac{MRAV}-1};
        \node[draw=redRSE, fill=redRSE, circle, inner sep=0.75pt] at (axis cs:\lobescale*1.05,0,0) {};
        \node[draw=black, fill=black, circle, inner sep=0.75pt] at (axis cs:0,0,0) {};
        \node[black, anchor=north east, font=\scriptsize] at (axis cs:0.1,0.05,0) {$O_{B_2}$};

        
        \end{axis}
        
        \begin{axis}[
          colormap/viridis,
          colorbar,
          colorbar style={
            ylabel={Normalized gain}, ylabel style={font=\scriptsize},
            ticklabel style={font=\scriptsize},
            width=0.18cm,
          },
          at={(4.0cm,0)}, anchor=center,
          width=5.8cm, height=4.8cm,
          axis lines=none, view={35}{25}, z buffer=sort,
        ]
        \DipoleSurf{\lobescale}

        \addplot3[thick, color=black] coordinates {(0,0,-1.4) (0,0,-1.0)};
        \addplot3[dashed, color=black] coordinates {(0,0,-0.8) (0,0,0.6)};
        \addplot3[-latex, solid, color=black] coordinates {(0,0,0.6) (0,0,1.2)};
        \node[anchor=west, text=black, font=\scriptsize] at (axis cs:0,0,1.2) {$\zz_{B_2}$};  

        \node[anchor=west, text=black, font=\scriptsize] at (axis cs:-0.30,0,-2.2) {(b)};
        
        \pgfmathsetmacro{\rxX}{cos(\tilt)}
        \pgfmathsetmacro{\rxZ}{sin(\tilt)}
        \addplot3[-latex, color=redRSE]
          coordinates {(0,0,0) (\lobescale*1.02*\rxX,0,\lobescale*1.02*\rxZ)};
        \node[redRSE, anchor=west, font=\scriptsize] at (axis cs:\lobescale*1.05*\rxX,0,\lobescale*1.05*\rxZ) {\ac{MRAV}-1};
        \node[draw=redRSE, fill=redRSE, circle, inner sep=0.75pt] at (axis cs:\lobescale*1.05*\rxX,0,\lobescale*1.05*\rxZ) {};
        \node[draw=black, fill=black, circle, inner sep=0.75pt] at (axis cs:0,0,0) {};
        \node[black, anchor=north east, font=\scriptsize] at (axis cs:0.1,0.05,0) {$O_{B_2}$};
        
        \addplot3[dashdotdotted, color=redRSE, domain=0:\tilt, samples=40]
          ({0.50*cos(x)},{0},{0.50*sin(x)});
        \node[redRSE, anchor=north west, font=\scriptsize] at (axis cs:0.45,0,0.57) {$\chi$};

        \addplot3[-latex, dashed, color=black] coordinates {(0,0,0) (1,0,0)};
        \node[anchor=west, text=black, font=\scriptsize] at (axis cs:1.0,0,0) {$\xx_{B_2}$};
        \end{axis}
    
    \end{tikzpicture}
    \vspace*{-1.45em}
    \caption{Normalized radiation pattern of a half-wave dipole antenna mounted along the body-frame $\zz_B$-axis. (a) \ac{MRAV}-1 aligned with the maximum-gain region in the equatorial plane. (b) \ac{MRAV}-1 tilted by an angle $\chi$, resulting in reduced gain. Colors indicate the normalized antenna gain.}
    \label{fig:doughnutShape}
    \vspace*{-1.35em}
\end{figure}
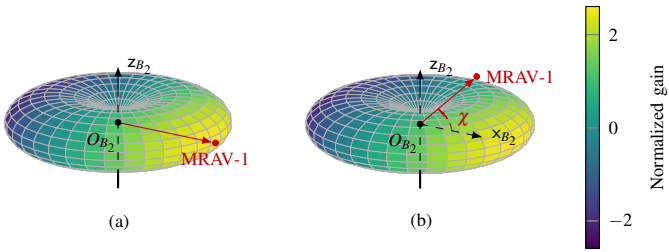

Coordination under jamming conditions is supported by an \ac{OCC} system, such as the \acl{UVDAR} platform \cite{HorynaICUAS2022, BonillaUVDAR2023}, which provides a jamming-immune, low-latency channel for \ac{MRAV}-to-\ac{MRAV} communication. Despite its limited bit rate, \ac{OCC} offers a wide field of view and transmits motion intent from \ac{MRAV}-2 to \ac{MRAV}-1, supporting predictive, coordinated \acs{NMPC} on \ac{MRAV}-1. \ac{OCC} therefore complements rather than replaces \ac{RF}, since it cannot provide the range or data rate required for mission communication.

Each aerial vehicle carries a single half-wave dipole antenna\footnote{The proposed approach does not rely on a specific parametric antenna model; it only requires a \textit{doughnut-shaped} radiation pattern with peak broadside gain. It therefore applies to other antennas with similar directional characteristics, such as those shown in Figure \ref{fig:doughnutShape}.} rigidly mounted along the body $\zz_B$-axis of frame $\pazocal{F}_{B_i}$, with $i = \{1, 2\}$ (see Figure \ref{fig:relayScenario}). These antennas exhibit a \textit{doughnut-shaped} radiation pattern \cite{ChenVTCF2018} whose inertial orientation is determined by vehicle attitude \cite{BonillaEUSIPCO21}: maximum gain occurs in the equatorial plane orthogonal to $\zz_B$, and the gain function $G(\cdot)$ (see Section~\ref{sec:modeling:CommunicationsChannel}) decays with the misalignment angle $\chi$, as illustrated in Figure~\ref{fig:doughnutShape}. Communication performance is therefore highly sensitive to attitude, especially during agile maneuvers. To isolate this coupling, we assume free-space \ac{LoS} propagation without small-scale fading, a realistic approximation for high-altitude and unobstructed aerial networks \cite{Shakeri2019IEEECST, Mozaffari2019CST}.  

End-to-end communication quality is constrained by the weaker of the two \ac{RF} links---from \ac{MRAV}-2 to relay or relay to \ac{BS}---so overall performance is determined by the minimum \ac{SINR};
accurate modeling of antenna orientation and motion-induced misalignment is therefore critical to maintain reliable connectivity \cite{BonillaEUSIPCO21}.

A ground-based jammer is assumed to emit persistent interference across the operational \ac{RF} bands, modeled as an isotropic radiator with uniform power distribution \cite{Mozaffari2019CST}, simultaneously degrading both links. Its location is initially unknown but becomes available after a short delay $\tau_E$ from jamming onset, consistent with standard localization techniques such as \ac{RSS}, \ac{TDoA}, or \ac{AoA} \cite{Bhamidipati2019ITJ, Juhlin2021EUSIPCO}. 



\section{Preliminaries}
\label{sec:preliminaries}

This section introduces the models and assumptions underlying the proposed approach. Table~\ref{tab:tableOfNotation} summarizes the notation used throughout the paper.



\subsection{System dynamics}
\label{sec:systemDynamics}

The \ac{MRAV} is modeled as a \ac{GTMR} system \cite{Ryll2019IJRR, HamandiIJRR2021}, actuated by $N_p$ motor-propeller units arbitrarily placed and oriented \ac{wrt} the body; the actuation layout determines whether the platform is coplanar or tilted \cite{Ryll2019IJRR, HamandiIJRR2021}.

 \begin{figure}[tb]
     \vspace*{-0.70em}
     \centering
     \scalebox{0.825}{
     \begin{tikzpicture}[scale=2.0, line cap=round, line join=round, >=Triangle]
         \draw[color=black!60, rotate around={-110:(0,0)}, fill=darkgreen!80, line width=1pt] (0,0) ellipse (0.95cm and 1.50cm);
         \draw[color=black!60, rotate around={-20:(0,0)}, fill=lightgreen!80, dashed, line width=1.5pt] (0,0) ellipse (1.5cm and 0.75cm);

         \fill (0,0) -- ++(0.2em,0) arc [start angle=0, end angle=90, radius=0.2em] -- ++(0,-0.4em) arc [start angle=270, end angle=180, radius=0.2em];
         \draw (0,0) [radius=0.2em] circle;

         \draw (-0.05,0) node[left]{$O_B$}; 
         \draw [-latex] (0,0) -- ({0.5*cos(65)},{0.5*sin(65)}) node[left]{$\zz_B$}; 
         \draw [-latex] (0,0) -- ({0.5*cos(20)},{0.5*sin(20)}) node[above]{$\yy_B$}; 
         \draw [-latex] (0,0) -- ({0.5*cos(-25)},{0.5*sin(-25)}) node[below]{$\xx_B$}; 

         \draw (-1.5,-1.5) node[left]{$O_W$}; 
         \node (circle) at (-1.5,-1.5) [circle, draw, scale=0.2, fill=black] {};
         \draw [-latex] (-1.5,-1.5) -- (-1.5,-1.0) node[left]{$\zz_W$}; 
         \draw [-latex] (-1.5,-1.5) -- (-1.15,-1.25) node[right]{$\yy_W$}; 
         \draw [-latex] (-1.5,-1.5) -- (-1.0,-1.5) node[below]{$\xx_W$}; 

         \draw[-latex, redRSE] (-1.5,-1.5) -- node[above]{$\mathbf{p}$} (0,0);
         \draw[-latex, redRSE] (-1.0,-1.75) to [out=-20, in=-70] node[left]{$\mathbf{R}$} (0.15, -0.15);

         \tikzset
         {%
           pics/cylinder/.style n args={3}{
             code={%
               \draw[pic actions] (135:#1) arc (135:315:#1) --++ (0,0,#2) arc (315:135:#1) -- cycle;
               \draw[pic actions] (0,0,#2) circle (#1);
               \foreach\z in {0,1}
               {
                 \begin{scope}[canvas is xy plane at z=\z*\h]
                   \coordinate (-cen\z) at       (0,0);
                   \coordinate (-ESE\z) at    (-#3:#1);
                   \coordinate (-ENE\z) at     (#3:#1);
                   \coordinate (-NNE\z) at  (90-#3:#1);
                   \coordinate (-NNW\z) at  (90+#3:#1);
                   \coordinate (-WNW\z) at (180-#3:#1);
                   \coordinate (-WSW\z) at (180+#3:#1);
                   \coordinate (-SSW\z) at (270-#3:#1);
                   \coordinate (-SSE\z) at (270+#3:#1);
                 \end{scope}
               }
             }},
         }
         \def\l{2}    
         \def\R{1}    
         \def\r{0.15} 
         \def\h{0.3}  
         \pic[fill=gray!30, rotate=-15, shift={(0,0,-\h)}] at (1.25,0.10) {cylinder={\r}{2*\h}{45}};
         \draw[-latex] (1.25,0.1225) -- (1.55,0.30) node[above]{$\zz_P$};
         \node at (1.1,-0.225) [above]{$\mathbf{p}_{m_1}$};
         \begin{scope}[shift={(1.405,0.25)},rotate around z=120,canvas is xy plane at z=\h]
           \draw[fill=gray!30] (0.0,0) sin  (0.25,0.1) cos  (0.65,0) sin  (0.25,-0.1) cos (0,0)sin (-0.25,0.1) cos (-0.65,0) sin (-0.25,-0.1) cos (0,0);
           \fill (0,0) circle (0.05);
           \draw[>-] (0.115,-0.175) arc (020:160:0.1) node [below] {$\tau$};
         \end{scope}
         \pic[fill=gray!30, rotate=175, shift={(0,0,-\h)}] at (-1.25,-0.10) {cylinder={\r}{2*\h}{45}};
         \draw[-latex] (-1.275,-0.11) -- (-1.545,-0.345) node[above]{$\zz_P$};
         \node at (1.0,-0.70) [above]{$\mathbf{p}_{m_2}$};
         \begin{scope}[shift={(-1.18,-0.025)},rotate around z=130,canvas is xy plane at z=\h]
           \draw[fill=gray!30] (0.0,0) sin  (0.25,0.1) cos  (0.65,0) sin  (0.25,-0.1) cos (0,0)sin (-0.25,0.1) cos (-0.65,0) sin (-0.25,-0.1) cos (0,0);
           \draw[->] (0.1,0.075) arc (020:160:0.1) node [below] {$\tau$};
           \fill (0,0) circle (0.05);
         \end{scope}
         \pic[fill=gray!30, rotate=70, shift={(0,0,-\h)}] at (-0.835,1.035) {cylinder={\r}{2*\h}{45}};
         \draw[-latex] (-0.845,1.045) -- (-0.965,1.335) node[right]{$\zz_P$};
         \node at (-0.95,-0.25) [above]{$\mathbf{p}_{m_3}$};
         \begin{scope}[shift={(-0.75,1.20)},rotate around z=20,canvas is xy plane at z=\h]
           \draw[fill=gray!30] (0.0,0) sin  (0.25,0.1) cos  (0.65,0) sin  (0.25,-0.1) cos (0,0)sin (-0.25,0.1) cos (-0.65,0) sin (-0.25,-0.1) cos (0,0);
           \draw[>-] (0.075,0.075) arc (00:140:0.1) node [left] {$\tau$};
           \fill (0,0) circle (0.05);
         \end{scope}
         \pic[fill=gray!30, rotate=240, shift={(0,0,-\h)}] at (0.85,-0.80) {cylinder={\r}{2*\h}{45}};
         \draw[-latex] (0.865,-0.85) -- (0.95,-1.10) node[right]{$\zz_P$};
         \node at (-0.825,0.625) [above]{$\mathbf{p}_{m_4}$};
         \begin{scope}[shift={(0.985,-0.74)},rotate around z=15,canvas is xy plane at z=\h]
           \draw[fill=gray!30] (0.0,0) sin  (0.25,0.1) cos  (0.65,0) sin  (0.25,-0.1) cos (0,0)sin (-0.25,0.1) cos (-0.65,0) sin (-0.25,-0.1) cos (0,0);
           \draw[->] (0.1,-0.2) arc (00:140:0.1) node [below left] {$\tau$};
           \fill (0,0) circle (0.05);
         \end{scope}
     \end{tikzpicture}
     } 
     \vspace*{-1.55em}
     \caption{A \ac{GTMR} system with world frame $\pazocal{F}_W$ and body frame $\pazocal{F}_B$.}
     \label{fig:systemModeling}
     \vspace*{-1.35em}
 \end{figure}

Two reference frames are used: the world frame $\pazocal{F}_W$ and the body frame $\pazocal{F}_B$, whose origin is attached to the vehicle's \ac{CoM}, as depicted in Figure \ref{fig:systemModeling}. The state $\mathbf{x}$ comprises position $\mathbf{p} \in \mathbb{R}^3$ and linear velocity $\mathbf{v} = \dot{\mathbf{p}} \in \mathbb{R}^3$ (both in $\pazocal{F}_W$), orientation $\bm{\eta}=(\varphi, \vartheta, \psi)^\top \in \mathbb{R}^3$ (Euler angles), and angular velocity $\bm{\omega} \in \mathbb{R}^3$ (in $\pazocal{F}_B$).
Motor $i \in \{1, \dots, N_p\}$ delivers thrust $\xi_i = c_{\xi_i} \Omega_i \zz_{P_i}$ and torque $\tau_i = \big( c_{f_i} \mathbf{p}_{m_i} \times \zz_{P_i} + c_{\tau_i} \zz_{P_i} \big) \Omega_i$, where $c_{\xi_i}$ and $c_{\tau_i}$ are thrust and torque coefficients, $\mathbf{p}_{m_i} \in \mathbb{R}^3$ is the motor position in $\pazocal{F}_B$, $\zz_{P_i} $ its axis, and $\Omega_i\ge 0$ the (squared) motor speed \cite{Ryll2019IJRR, HamandiIJRR2021}. Collecting inputs as $\mathbf{u}=\bm{\xi}=(\xi_1, \dots, \xi_{N_p})^\top$, the Newton–-Euler dynamics read:
\begin{equation}\label{eq:multirotorDynamics}
\left\{
    \begin{array}{l}
    \dot{\mathbf{p}} = \mathbf{v} \\
    \dot{\bm{\eta}} = \mathbf{T}(\bm{\eta}) \bm{\omega} \\
    m \dot{\mathbf{v}} = -mg \mathbf{e}_3 + \mathbf{R}(\bm{\eta})
        \mathbf{F} \mathbf{u} \\
    \mathbf{J} \dot{\bm{\omega}} = - \bm{\omega} \times \mathbf{J}
        \bm{\omega} + \mathbf{M} \mathbf{u}
	\end{array}
\right.,
\end{equation}
where $m$ is the mass, $g$ the gravitational constant, $\mathbf{e}_3=(0,0,1)^\top$, $\mathbf{J}\in\mathbb{R}^{3\times 3}$ the inertia matrix in $\pazocal{F}_B$, $\mathbf{R}(\bm{\eta}) \in \mathbb{R}^{3 \times 3}$ the rotation from $\pazocal{F}_B$ to $\pazocal{F}_W$, $\mathbf{T}(\bm{\eta}) \in \mathbb{R}^{3 \times 3}$ the Euler-rate map, and $\mathbf{F}, \mathbf{M} \in \mathbb{R}^{3\times N_p}$ the allocation matrices mapping motor thrusts to net force and moment at the \ac{CoM} \cite{Ryll2019IJRR, HamandiIJRR2021}.

To capture actuator bandwidth limits, the commanded input is the time derivative of motor speeds, $\dot{\mathbf{u}} = \dot{\bm{\Omega}} \in \mathbb{R}^{N_p}$, and the state is augmented as $\bar{\mathbf{x}} = (\mathbf{p}^\top, \bm{\eta}^\top, \mathbf{v}^\top, \bm{\omega}^\top, \mathbf{u}^\top)^\top \in \mathbb{R}^{12+N_p}$. 
This formulation enables explicit actuator constraints:
\begin{equation}
    \underline{\gamma} \leq \mathbf{u} \leq \bar{\gamma} \quad
    \mathrm{and} \quad
    \dot{\underline{\gamma}} \leq \dot{\mathbf{u}} \leq \dot{\bar{\gamma}},
\end{equation}
where $\underline{\gamma},\bar{\gamma}$ are minimum/maximum motor speeds and $\dot{\underline{\gamma}}, \dot{\bar{\gamma}}$ their acceleration bounds (possibly speed-dependent). A base-station body frame $\pazocal{F}_{B_0}$ is also defined at $\mathbf{p}_0$, aligned with $\pazocal{F}_W$, for notational convenience.

\begin{table}[t]
    \centering
    \caption{Summary of notation.}
    \vspace{-1.0em}
    \label{tab:tableOfNotation}
    \includegraphics[width=\columnwidth]{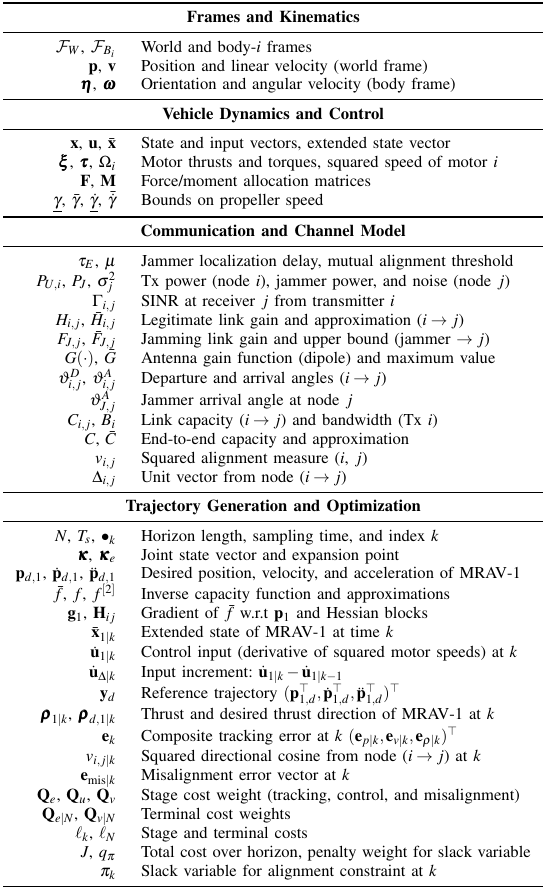}
    \vspace{-1.35em}
\end{table}



\subsection{Communication channel model}
\label{sec:modeling:CommunicationsChannel}

The \ac{RF} links are modeled using the free-space path loss model under unobstructed \ac{LoS} conditions \cite{SilanoSMC2025}; small-scale fading and shadowing are neglected, consistent with aerial scenarios where vehicles operate at high altitude with a dominant \ac{LoS} component \cite{Shakeri2019IEEECST, Mozaffari2019CST}. Each \ac{MRAV} carries a half-wave dipole antenna rigidly mounted along its body-frame $\zz_B$-axis (see Figure \ref{fig:relayScenario}), coupling the antenna's orientation in $\pazocal{F}_W$ directly to vehicle attitude $\bm{\eta}$ and making link performance sensitive to motion-induced misalignment. 

For transmitter node $i$ and receiver node $j$, the received \ac{SINR} is:
\begin{equation} \label{eq:SINR}
    \Gamma_{i,j} = \frac{H_{i,j}^2 P_{U,i}}{F_{J,j}^2 P_{J} + \sigma_j^2},
\end{equation}
where $P_{U,i}$ is the transmit power of node $i$, $P_J$ the jammer power, and $\sigma^2_j$ the receiver noise at node $j$. 

The legitimate link gain from transmitter $i$ to receiver $j$ is:
\begin{equation} \label{eq:channelGain}   
     H_{i,j} = \frac{\sqrt{G(\vartheta_{i,j}^{D})\,G(\vartheta_{i,j}^{A})}}
                   {\|\mathbf{p}_{j} - \mathbf{p}_{i}\|}, 
\end{equation}
where $\vartheta_{i, j}^{D}$ and $\vartheta_{i,j}^{A}$ are the departure and arrival angles of the \ac{RF} link \ac{wrt} the transmitting and receiving antenna boresights, respectively, Similarly, the jamming link gain from jammer $J$ to node $j$ is:
\begin{equation} \label{eq:jammerGain}   
    F_{J,j} = \frac{\sqrt{G(\vartheta^A_{J,j})}}
                   {\|\mathbf{p}_{J}-\mathbf{p}_j\|},
\end{equation}
where $\vartheta^A_{J,j}$ is the \ac{AoA} of the jamming signal at node $j$ relative to its receiving antenna orientation. 

In both~\eqref{eq:channelGain} and~\eqref{eq:jammerGain}, $G(\cdot)$ denotes the directional antenna gain determined by angular misalignment between the incident signal and the boresight, which is highly nonlinear \ac{wrt} the \ac{AoA} even for simple antennas \cite{BonillaEUSIPCO21}; the \ac{AoA} itself is highly nonlinear \ac{wrt} the relative position and orientation of both \acp{MRAV} (see Appendix~\ref{sec:antennaRadiation}).
The link geometry is fully determined by positions $\mathbf{p}_i$, $\mathbf{p}_{j}$, and $\mathbf{p}_J$; angle definitions and the antenna radiation model are detailed in Appendix~\ref{sec:antennaRadiation}. 

The instantaneous capacity of the legitimate \ac{RF} link from $i$ to $j$ is:
\begin{equation} \label{eq:capacityRF}
    C_{i,j} = B_{i}\, \log_2\left(1+\Gamma_{i,j}\right),
\end{equation} 
where $B_{i}$ is the bandwidth allocated to transmitter $i$. 



\section{Modular Control Architecture}
\label{sec:modularControlFramework}

This section introduces a modular control architecture for relay \ac{MRAV}-1. Assuming dynamic bandwidth allocation and equalizing the capacity of both links---since the end-to-end rate is limited by the weaker link \cite{Laneman2004TIT, Goldsmith2005}---the equivalent end-to-end rate is:
\begin{equation}\label{eq:balancedCapacity}
    C = \frac{B \log_2(1 + \Gamma_{2,1}) \log_2(1 + \Gamma_{1,0})}
             {\log_2(1 + \Gamma_{2,1}) + \log_2(1 + \Gamma_{1,0})},
\end{equation}
where $\Gamma_{2,1}$ and $\Gamma_{1,0}$ denote the \ac{SINR} of the source-to-relay and relay-to-\ac{BS} links, respectively.

In~\eqref{eq:balancedCapacity}, the end-to-end rate is dominated by the smaller of the two link \acp{SINR}---referred to as the \emph{weakest link}---since doubling the capacity of the stronger link yields a negligible improvement in $C$, while even a small degradation of the weaker link substantially reduces it. The trajectory generator (Section~\ref{sec:trajectoryGenerator}) is designed to strengthen the weakest link via the following $\max$--$\min$ formulation.

Using index $0$ for the \ac{BS} and indices $1$, $2$ for \ac{MRAV}-1 and \ac{MRAV}-2, the ideal control problem is:
\begin{equation} \label{eq:optimizationProblemSINR}
    \begin{split}
    &\maximize_{\bar{\mathbf{x}}_1, \dot{\mathbf{u}}_1}\ \
    \displaystyle C\left(\Gamma_{2,1}(\bar{\mathbf{x}}_1),
        \Gamma_{1,0}(\bar{\mathbf{x}}_1)\right) \\
    &\quad \text{s.t.} \quad \text{eq.}~\eqref{eq:multirotorDynamics}, \\
    &\qquad \quad\;\;
        \underline{\gamma} \leq \mathbf{u}_1 \leq \bar{\gamma},\;
        \dot{\underline{\gamma}} \leq \dot{\mathbf{u}}_1 \leq
        \dot{\bar{\gamma}}, \\
    &\qquad \quad\;\; \text{with}\ \Gamma_{i,j}\ \text{given
        by}~\eqref{eq:SINR},\
        (i,j)\in\{(2,1),(1,0)\}.
    \end{split}\, .
\end{equation}

Because $\Gamma_{i,j}$ depends nonlinearly on vehicle position and orientation through directional antenna gains (see \eqref{eq:channelGain}--\eqref{eq:jammerGain}), solving~\eqref{eq:optimizationProblemSINR} in real time is intractable, so we adopt conservative approximations detailed below.

The approach consists of two integrated modules. The first is a high-level trajectory generator that solves a $\max$–$\min$ problem to strengthen the weakest \ac{RF} link under interference (see Section \ref{sec:trajectoryGenerator}), incorporating time-varying channel conditions, antenna directionality, and jammer position to produce communications-aware motion references that feed the second module. The latter is a low-level \acs{NMPC} that tracks the references while enforcing vehicle dynamics and actuator constraints (see Section \ref{sec:communicationAwareNMPC}), enforcing directional alignment through geometric misalignment metrics and exploiting motion intent shared via \ac{OCC} for coordination.



\subsection{Trajectory generator}
\label{sec:trajectoryGenerator}

The trajectory generator produces references for \ac{MRAV}-1 to improve network connectivity under jamming by repositioning the relay to maximize the end-to-end capacity between \ac{MRAV}-2 and the ground \ac{BS}.

\subsubsection{Jammer and legitimate channels simplification}
To remove nonlinearities from antenna directionality, the jamming link is upper-bounded as
\begin{equation}\label{eq:barFJj}
    \bar{F}_{J,j} = \frac{\sqrt{\bar{G}}}{\lVert \mathbf{p}_J -
    \mathbf{p}_j \rVert}, \quad j\in\{1,0\},
\end{equation}
where $\bar{G} \triangleq \max_\vartheta \, G(\vartheta)$ is the maximum antenna gain, representing the worst case condition in which the jammer lies in the maximum-gain (equatorial) plane of the receiving antenna and providing a conservative upper bound on interference power \ac{wrt} \ac{MRAV}-1 orientation.

The isotropic-jammer model in~\eqref{eq:barFJj} is conservative \ac{wrt} directional jammers of equal total radiated power. For any direction-dependent jammer with peak gain $G_J(\cdot)\le \bar G_J$ and total radiated power $P_J$, the interference at receiver $j$ satisfies $F_{J,j}^{\mathrm{dir}} \le F_{J,j}^{\mathrm{iso}} = \sqrt{\bar{G}}/\|\mathbf{p}_J - \mathbf{p}_j\|$ whenever the jammer's broadside is not pointed at $j$; when it is, the bound~\eqref{eq:barFJj} is recovered using the receiver-side antenna gain upper bound $\bar G$.

Similarly, the legitimate channel gain is conservatively approximated as
\begin{equation}
    \bar{H}_{i,j} = \frac{\sqrt{\mu}\,\bar{G}}{\lVert \mathbf{p}_j -
    \mathbf{p}_{i} \rVert}, \quad (i,j)\in\{(2,1),(1,0)\},
\end{equation}
yielding a bounded \ac{SINR} $\bar{\Gamma}_{i,j}$ that captures range and alignment effects while decoupling antenna gains from the \acs{NMPC} (see Section~\ref{sec:communicationAwareNMPC}). Substituting $\bar{\Gamma}_{i,j}$ into~\eqref{eq:balancedCapacity} gives the conservative end-to-end capacity:
\begin{equation} \label{eq:revisedCapacityFcn}
    \bar{C} = \frac{B \log_2(1 + \bar{\Gamma}_{2,1})\,
                      \log_2(1 + \bar{\Gamma}_{1,0})}
                   {\log_2(1 + \bar{\Gamma}_{2,1}) +
                    \log_2(1 + \bar{\Gamma}_{1,0})}.
\end{equation}

\subsubsection{Quadratic cost function construction}
To obtain a quadratic-programming-compatible cost function for real-time optimization, we define the inverse capacity function:
\begin{equation}
    f(\bm{\kappa}, P_J) = \frac{1}{B} \left(
    \frac{1}{\log_2(1 + \bar{\Gamma}_{2,1})} +
    \frac{1}{\log_2(1 + \bar{\Gamma}_{1,0})} \right),
\end{equation}
where $\bm{\kappa} \triangleq (\mathbf{p}_1^\top, \mathbf{p}_2^\top)^\top$ collects the relay and source positions, with $\mathbf{p}_2$ available through the \ac{OCC} link.

Since $f$ is nonlinear and non-convex, it is locally approximated by a second-order Taylor expansion around $\bm{\kappa}_e$ (see Appendix~\ref{sec:taylorExpansion}):
\begin{equation}
\begin{split}
    f^{[2]}(\bm{\kappa}; \bm{\kappa}_e, P_J) &= \bar{f}(\bm{\kappa}_e,
    P_J) + \nabla_{\bm{\kappa}} \bar{f}(\bm{\kappa}_e, P_J)^\top
    (\bm{\kappa} - \bm{\kappa}_e) \\
    &\quad + \tfrac{1}{2}(\bm{\kappa} - \bm{\kappa}_e)^\top
    \nabla^2_{\bm{\kappa}} \bar{f}(\bm{\kappa}_e, P_J)
    (\bm{\kappa} - \bm{\kappa}_e),
\end{split}
\end{equation}
where $\bar{f}$ denotes $f$ under the conservative channel model. The expansion point $\bm{\kappa}_e$ is updated from previous solutions, ensuring consistency across iterations. The Taylor expansion is taken only \ac{wrt} positions in $\bm{\kappa}$; angular dependence of $G(\cdot)$ is instead enforced in the \acs{NMPC} (see~\eqref{eq:mutualMisalignment}).

Treating the source motion as exogenous and focusing on the relay position $\mathbf{p}_1$, the resulting quadratic model is:
\begin{equation}
    h(\mathbf{p}_1) = \mathbf{g}_1^\top \bm{\delta}_1 +
    \tfrac{1}{2} \bm{\delta}_1^\top \mathbf{H}_{11} \bm{\delta}_1 +
    \bm{\delta}_1^\top \mathbf{H}_{12} \bm{\delta}_2,
\end{equation}
with $\bm{\delta}_i = \mathbf{p}_i - \mathbf{p}_{i_e}$, where $\mathbf{g}_1$ is the gradient of $\bar{f}$ \ac{wrt} $\mathbf{p}_1$ and $\mathbf{H}_{ij}$ are the Hessian blocks (see Appendix~\ref{sec:taylorExpansion}).

Minimizing $h$ \ac{wrt} $\mathbf{p}_1$ yields the relay references:
\begin{align}
    \mathbf{p}_{d,1} &= \mathbf{p}_{1_e} - \mathbf{H}_{11}^{-1}
        (\mathbf{g}_1 + \mathbf{H}_{12} \bm{\delta}_2),
        \label{eq:references_position} \\
    \dot{\mathbf{p}}_{d,1} &= -\mathbf{H}_{11}^{-1} \mathbf{H}_{12}\,
        \dot{\mathbf{p}}_2^\top, \label{eq:references_velocity} \\
    \ddot{\mathbf{p}}_{d,1} &= -\mathbf{H}_{11}^{-1} \mathbf{H}_{12}\,
        \ddot{\mathbf{p}}_2^\top. \label{eq:references_acceleration}
\end{align}

This reformulation replaces explicit radiation-pattern models with geometric alignment constraints, yielding a tractable optimization that preserves the dominant dependence on range, antenna orientation (through $\ddot{\mathbf{p}}_{d,1}$, as clarified in Section~\ref{sec:communicationAwareNMPC}), and jammer interference.

\subsubsection{Regularization for invertibility}
Computing $\mathbf{p}_{d,1}$ requires $\mathbf{H}_{11}$ to be invertible, which is generally satisfied but may fail near flat regions or saddle points of the approximated capacity. As a safeguard, if $\mathbf{H}_{11}$ is near-singular, the expansion point is perturbed along the local gradient of $\bar{f}$:
\begin{equation}
    \bm{\kappa}_e^\prime = \bm{\kappa}_e + \varepsilon\bm{\kappa},
    \qquad \lVert \varepsilon\bm{\kappa} \rVert \ll 1,
\end{equation}
with $\varepsilon\bm{\kappa}$ chosen to restore positive definiteness. In practice, such regularization is rarely needed, as successive \acs{NMPC} solutions typically lie in well-conditioned regions of the cost landscape.

\subsubsection{Approximation properties}
The trajectory generator involves two approximation layers, characterized below.

\textit{Conservativeness of the channel surrogate.} By construction $\bar{F}_{J,j} \ge F_{J,j}$ \eqref{eq:barFJj}, with equality when the jammer lies in the broadside plane of the receiving antenna. For the legitimate channel, $\bar{H}_{i,j} = \sqrt{\mu}\,\bar{G}/\|\mathbf{p}_j - \mathbf{p}_i\|$ is conservative whenever $G(\vartheta^D_{i,j}) G(\vartheta^A_{i,j}) \ge \mu\bar{G}^2$, which holds for the doughnut-shaped dipole pattern provided both link ends maintain joint alignment of at least
$\mu$. Under this condition, $\bar{H}_{i,j} \le H_{i,j}$, $\bar{\Gamma}_{i,j} \le \Gamma_{i,j}$, and $\bar{C} \le C$. Section~\ref{sec:communicationAwareNMPC} shows how the \acs{NMPC} enforces precisely this alignment condition at runtime, closing the loop between the conservative design and its realized guarantee.

\textit{Taylor remainder bound.} Let $\pazocal{D} \subset \mathbb{R}^6$ be a compact operating set excluding $\mathbf{p}_j = \mathbf{p}_i$ and $\bar{\Gamma}_{i,j} \le 0$, neither of which arises in the mission. On $\pazocal{D}$, $\bar{f}(\cdot, P_J)$ is three times continuously differentiable. By Lagrange's remainder theorem:
\begin{equation}\label{eq:taylorRemainder}
    \bigl|\bar{f}(\bm{\kappa}, P_J) - f^{[2]}(\bm{\kappa}; \bm{\kappa}_e,
    P_J)\bigr| \;\le\; \frac{M_3}{6}\,\|\bm{\kappa} - \bm{\kappa}_e\|^3,
\end{equation}
where $M_3 = \sup_{\bm{\kappa}' \in \pazocal{D}} \|\nabla^3_{\bm{\kappa}} \bar{f}(\bm{\kappa}', P_J)\|$. Since $\bm{\kappa}_e$ is updated at every \acs{NMPC} step, $\|\bm{\kappa} - \bm{\kappa}_e\| \le \bar{v} T_s$. Formula~\eqref{eq:taylorRemainder} is evaluated numerically in Section~\ref{sec:nominalScenario}.



\subsection{Communications-aware NMPC}
\label{sec:communicationAwareNMPC}

This section presents the \acs{NMPC} scheme that tracks the reference trajectory from Section~\ref{sec:trajectoryGenerator} while enforcing vehicle dynamics, actuator limits, and antenna-alignment constraints in a receding-horizon manner.

\subsubsection{Theoretical scope}
Recursive feasibility and asymptotic stability are not claimed: problem~\eqref{eq:NMPC_formulation} omits terminal ingredients (terminal cost, terminal set, control Lyapunov function~\cite{Mayne2000Automatica}). Two design choices motivate this.
First, the alignment constraint~\eqref{eq:nmpc_alignment_constraint} is posed as a soft constraint via slack $\pi_k$ to ensure feasibility recovery under transient geometric configurations; terminal ingredients would conflict with this soft formulation. Second, the mission is naturally finite-horizon, and the relevant metric is the realized end-to-end capacity rather than asymptotic regulation. Empirical reliability, including feasibility metrics, is documented in Sections~\ref{sec:baselineComparison},~\ref{sec:fadingRobustness}, and~\ref{sec:robustnessModelUncertainty}.

\subsubsection{Legitimate channel simplification via misalignment constraints}
To prevent link degradation from antenna misalignment, we introduce the squared directional cosine:
\begin{equation}\label{eq:directionalCosine}
    v_{i,j} \triangleq \left( \langle \Delta_{i,j},
    \mathbf{R}_i(\bm{\eta}_i)\zz_{B_i} \rangle \right)^2,
\end{equation}
where $\Delta_{i,j} = (\mathbf{p}_j - \mathbf{p}_i)/\lVert \mathbf{p}_j - \mathbf{p}_i \rVert$ is the unit vector from $i$ to $j$,
$\mathbf{R}_i(\bm{\eta}_i)$ is the body-to-world rotation of node $i$, and $\zz_{B_i}$ is its body-frame boresight. The quantity $v_{i,j} \in [0,1]$ measures angular misalignment: $v_{i,j}=0$ at maximum gain (equatorial plane) and $v_{i,j}=1$ at complete misalignment (boresight-parallel link).

To ensure \ac{MRAV}-1 maintains sufficient alignment simultaneously toward \ac{MRAV}-2 and the ground \ac{BS}, we impose the mutual constraint:
\begin{equation} \label{eq:mutualMisalignment}
    (1 - v_{1,2})(1 - v_{1,0}) \geq \mu,
\end{equation}
where $\mu \in [0,1]$ sets the minimum joint alignment level. This enforces balanced antenna orientation across both links, consistent with end-to-end performance being bottlenecked by the weakest one. 

\subsubsection{Discrete-time formulation}
The system is discretized over a prediction horizon of $N$ steps with sampling period $T_s>0$, defining the time grid $t_k = kT_s$,
$k\in\{0,\dots,N\}$; the subscript $k$ denotes the $k$th predicted value (e.g., $\mathbf{p}_{1|k}$, $\mathbf{v}_{1|k}$).

At prediction step $k$, the augmented state is $\bar{\mathbf{x}}_{1|k} = (\mathbf{p}_{1|k}^\top, \bm{\eta}_{1|k}^\top, \mathbf{v}_{1|k}^\top, \bm{\omega}_{1|k}^\top, \mathbf{u}_{1|k}^\top)^\top$ and the control input is $\dot{\mathbf{u}}_{1|k}$. \ac{MRAV}-2's predicted position, velocity, and acceleration $\mathbf{p}_{2|k}$, $\dot{\mathbf{p}}_{2|k}$, $\ddot{\mathbf{p}}_{2|k}$ are available to \ac{MRAV}-1 via \ac{OCC}, allowing the trajectory generator to evaluate \eqref{eq:references_position}--\eqref{eq:references_acceleration} at each $k$ and supply the discrete-time reference $\mathbf{y}_{d|k} = (\mathbf{p}_{d,1|k}^\top, \dot{\mathbf{p}}_{d,1|k}^\top, \ddot{\mathbf{p}}_{d,1|k}^\top)^\top$. 

\subsubsection{Objective function}
The \acs{NMPC} minimizes three objectives---reference tracking, smooth actuation, and antenna alignment---encoded in stage and terminal costs. The composite tracking error is:
\begin{equation}\label{eq:errorReferenceTrajectory}    
    \mathbf{e}_k =
    \begin{pmatrix}
        \mathbf{e}_{p,k} \\
        \mathbf{e}_{v,k} \\
        \mathbf{e}_{\rho,k}
    \end{pmatrix} =
    \begin{pmatrix}
        \mathbf{p}_{1|k} - \mathbf{p}_{d,1|k} \\
        \mathbf{v}_{1|k} - \dot{\mathbf{p}}_{d,1|k} \\
        \bm{\rho}_{1|k} - \bm{\rho}_{d,1|k}
    \end{pmatrix},
\end{equation}
where $\bm{\rho}_{1|k} = \mathbf{R}_1(\bm{\eta}_{1|k})\zz_{B_1}$ is the thrust direction in $\pazocal{F}_W$ and $\bm{\rho}_{d,1|k} = -\ddot{\mathbf{p}}_{d,1|k}/\|\ddot{\mathbf{p}}_{d,1|k}\|$ is the desired unit thrust direction.

To discourage aggressive actuation, the input variation is penalized:
\begin{equation}\label{eq:outputVariationCost}
    \dot{\mathbf{u}}_{\Delta|k} = \dot{\mathbf{u}}_{1|k} -
    \dot{\mathbf{u}}_{1|k-1},
\end{equation}
where $\dot{\mathbf{u}}_{1|k-1}$ is the control input at the previous prediction step (initialized at $k=0$ with the last applied action).

The misalignment penalty vector for the two active links is:
\begin{equation}\label{eq:misalignmentCost}
    \mathbf{e}_{\mathrm{mis}|k} =
    \begin{pmatrix}
        v_{1,2|k} \\
        v_{1,0|k}
    \end{pmatrix},
\end{equation}
which directly penalizes directional degradation on both links since $v_{i,j} \in [0,1]$ increases with misalignment.

The stage cost is:
\begin{equation}
    \ell_k = \mathbf{e}_k^\top \mathbf{Q}_e\,\mathbf{e}_k +
    \dot{\mathbf{u}}_{\Delta|k}^\top \mathbf{Q}_u\,\dot{\mathbf{u}}_{\Delta|k}
    + \mathbf{e}_{\mathrm{mis}|k}^\top \mathbf{Q}_v\,\mathbf{e}_{\mathrm{mis}|k},
\end{equation}
where $\mathbf{Q}_e \succeq 0$, $\mathbf{Q}_u \succ 0$, $\mathbf{Q}_v \succeq 0$ tune tracking fidelity, control smoothness, and alignment, respectively. The terminal cost is:
\begin{equation}
    \ell_N = \mathbf{e}_N^\top\mathbf{Q}_{e|N}\,\mathbf{e}_N +
    \mathbf{e}_{\mathrm{mis}|N}^\top\mathbf{Q}_{v|N}\,\mathbf{e}_{\mathrm{mis}|N},
\end{equation}
with $\mathbf{Q}_{e|N} \succeq 0$, $\mathbf{Q}_{v|N} \succeq 0$. The total cost over the horizon is:
\begin{equation}
    J = \sum_{k=0}^{N-1} \ell_k + \ell_N.
\end{equation}

\subsubsection{Optimal control problem} 
The communications-aware \acs{NMPC} is posed as:

\vspace*{-1.25em}
\begin{subequations}\label{eq:NMPC_formulation}
    \small
    \begin{align}
    \minimize_{\bar{\mathbf{x}}_1, \dot{\mathbf{u}}_1, \pi_k} \quad
        & \sum_{k=0}^{N-1} \ell_k + \ell_N +
          \sum_{k=0}^{N} q_\pi\,\pi_k^2
          \label{eq:nmpc_cost} \\[0.2em]
    \text{s.t.} \quad
        & \bar{\mathbf{x}}_{1,0} = \bar{\mathbf{x}}_{1|k},
          \label{eq:nmpc_initial_condition} \\[0.2em]
        & \bar{\mathbf{x}}_{1|k+1} = \Phi(\bar{\mathbf{x}}_{1|k},
          \dot{\mathbf{u}}_{1|k}),\quad k = \{0,\dots,N-1\},
          \label{eq:nmpc_dynamics} \\[0.2em]
        & \underline{\gamma} \leq \mathbf{u}_{1|k} \leq \bar{\gamma},
          \quad k = \{0,\dots,N-1\},
          \label{eq:nmpc_input_bounds} \\[0.2em]
        & \dot{\underline{\gamma}} \leq \dot{\mathbf{u}}_{1|k} \leq
          \dot{\bar{\gamma}},\quad k = \{0,\dots,N-1\},
          \label{eq:nmpc_input_rate_bounds} \\[0.2em]
        & (1 - v_{1,2|k})(1 - v_{1,0|k}) \geq \mu - \pi_k,
          \label{eq:nmpc_alignment_constraint} \\[0.2em]
        & \pi_k \geq 0,\quad k = \{0,\dots,N\}.
          \label{eq:nmpc_slack_nonnegativity}
    \end{align}
    \normalsize
\end{subequations}

The state-transition map $\Phi(\cdot)$ in~\eqref{eq:nmpc_dynamics} corresponds to numerical integration of the continuous-time
model~\eqref{eq:multirotorDynamics} over one sampling interval; actuator saturation and rate limits are enforced
by~\eqref{eq:nmpc_input_bounds}--\eqref{eq:nmpc_input_rate_bounds}.

The alignment constraint~\eqref{eq:nmpc_alignment_constraint} promotes simultaneous antenna alignment across both communication links by coupling the alignment metrics $v_{1,2|k}$ and $v_{1,0|k}$. Their product captures the joint alignment quality, enabling the controller to adaptively distribute the alignment effort between the two links instead of satisfying independent per-link constraints, which could unnecessarily restrict the feasible set.

To preserve feasibility when the desired alignment cannot be achieved because of vehicle dynamics or unfavorable geometry, the constraint is relaxed using the non-negative slack variable $\pi_k$. The corresponding quadratic penalty, $q_\pi\pi_k^2$, discourages excessive violations while allowing controlled relaxation when necessary.



\section{Simulation Results}
\label{sec:simulationResults}

Numerical simulations were conducted in MATLAB using the MATMPC toolbox \cite{ChenECC2019}. The nonlinear optimal control problem was discretized via a fourth-order Runge--Kutta scheme with $T_s = \SI{15}{\milli\second}$ and solved using qpOASES \cite{Ferreau2014}, on a workstation with an Intel\textsuperscript{\textregistered} Core\textsuperscript{\texttrademark} Ultra i7-255H CPU (1.80~GHz) and 64~GB RAM running Ubuntu~24.04.



\subsection{Nominal scenario}
\label{sec:nominalScenario}

The simulation emulates a power-line inspection mission. \ac{MRAV}-2 follows a predefined trajectory adapted from \cite{SilanoRAL2021}, reaching the mid-height of the tower and then the insulator region near the top (${\approx}\SI{18}{\meter}$ high, $\SI{3}{\meter}$ wide), inducing significant variation in \ac{LoS} geometry. \ac{MRAV}-1 continuously repositions to maintain directional alignment with both \ac{MRAV}-2 and the \ac{BS}, counteracting a stationary isotropic jammer (localized at
$\tau_E=\SI{2}{\second}$) transmitting constant power. The simulated workspace spans $\SI{14}{\meter}\times\SI{12}{\meter}\times\SI{19}{\meter}$.

The platform is a tilted-propeller hexarotor enabling decoupled translation and rotation for accurate alignment under dynamic constraints. Parameters in Table~\ref{tab:controlParameters} are representative of a medium-scale research \ac{MRAV}. The prediction horizon is $T=\SI{0.45}{\second}$, discretized into $N=30$ shooting nodes. Control inputs are $\dot{\bm{\Omega}}$, updated at $\SI{500}{\hertz}$; the \ac{MRAV}-1 model (Section~\ref{sec:systemDynamics}) is integrated at $\SI{1}{\kilo\hertz}$ to
capture fast actuation dynamics.

\begin{table}[tb]
    \centering
    \caption{Optimization parameters.}
    \vspace{-1.0em}
    \label{tab:controlParameters}
    \begin{adjustbox}{max width=0.98\columnwidth}
    \begin{tabular}{c|c|c|c}
        \toprule
        \textbf{Sym.} & \textbf{Value} & \textbf{Sym.} & \textbf{Value} \\
        \midrule
        $\mathbf{p}_0$ & $[0,\,0,\,0]^\top\,\si{\meter}$ & $\bm{\eta}_0, \bm{\eta}_{1|0}, \bm{\eta}_{2|0}$ & $[0,\,0,\,0]^\top\,\si{\radian}$ \\
        $\mathbf{p}_{1|0}$ & $[-6.95,\,5.79,\,1.72]^\top\,\si{\meter}$ & $D$, $\bar{G}$ & $1.64$, $1$  \\
        $\mathbf{p}_{2|0}$ & $[-3.04,\,5.79,\,1.72]^\top\,\si{\meter}$ & $\mu$, $B$ & $0.2$, $1$ \\
        $\mathbf{p}_{J}$ & $[-6.95,\,-5.79,\,1.72]^\top\,\si{\meter}$ & $\bm{\eta}_{J}$ & $[0,\,0,\,0]^\top\,\si{\radian}$ \\
        $c_{\xi}$, $c_{\tau}$ & $1.18\!\times\!10^{-3}$,\;$2.5\!\times\!10^{-5}$ & $\mathbf{J}$ & $\mathrm{diag}(0.11,\,0.11,\,0.19)\,\si{\kilogram\meter^2}$ \\
        $T_s$, $N$ & $15\,\si{\milli\second}$,\;$30$ & $g$, $m$, $N_p$ & $9.81\,\si{\meter\per\square\second}$,\;$2.57\,\si{\kilogram}$, $6$ \\
        $\bar{\gamma}$,\;$\underline{\gamma}$ & $100,\;16\,\si{\hertz}$ & 
        $\dot{\bar{\gamma}}$, $\dot{\underline{\gamma}}$ & $400,\;-300\,\si{\hertz\per\second}$ \\
        $\mathbf{Q}_e$ & $\mathrm{diag}(0.1,\;0.1,\;0.1)$ & $\mathbf{Q}_v$, $\mathbf{Q}_{e|N}$ & $\mathrm{diag}(10,\,10)$, $\mathrm{diag}(0.1,\;0.1,\;0.1)$ \\
        $\mathbf{Q}_u$, $\mathbf{Q}_{v|N}$ & $\mathrm{diag}(10,\,10)$, $\mathrm{diag}(10,\dots,10)$ & $P_U$, $P_J$ & $1\,\si{\watt}, 1\,\si{\watt}$ \\
        $\sigma_J$, $q_\pi$ & $1\,\si{\sqrt{\watt}}$, $\num{1e4}$ & $\tau_E$, on-off & $2\,\si{\second}$, $5\,\si{\second}$ \\
        \bottomrule
    \end{tabular}
    \end{adjustbox}
    \vspace*{-1.35em}
\end{table}

\begin{figure}[tb]
    \centering
    \adjincludegraphics[width=0.74\columnwidth, trim={{.075\width} {.070\height} {.070\width} {.075\height}}, clip]{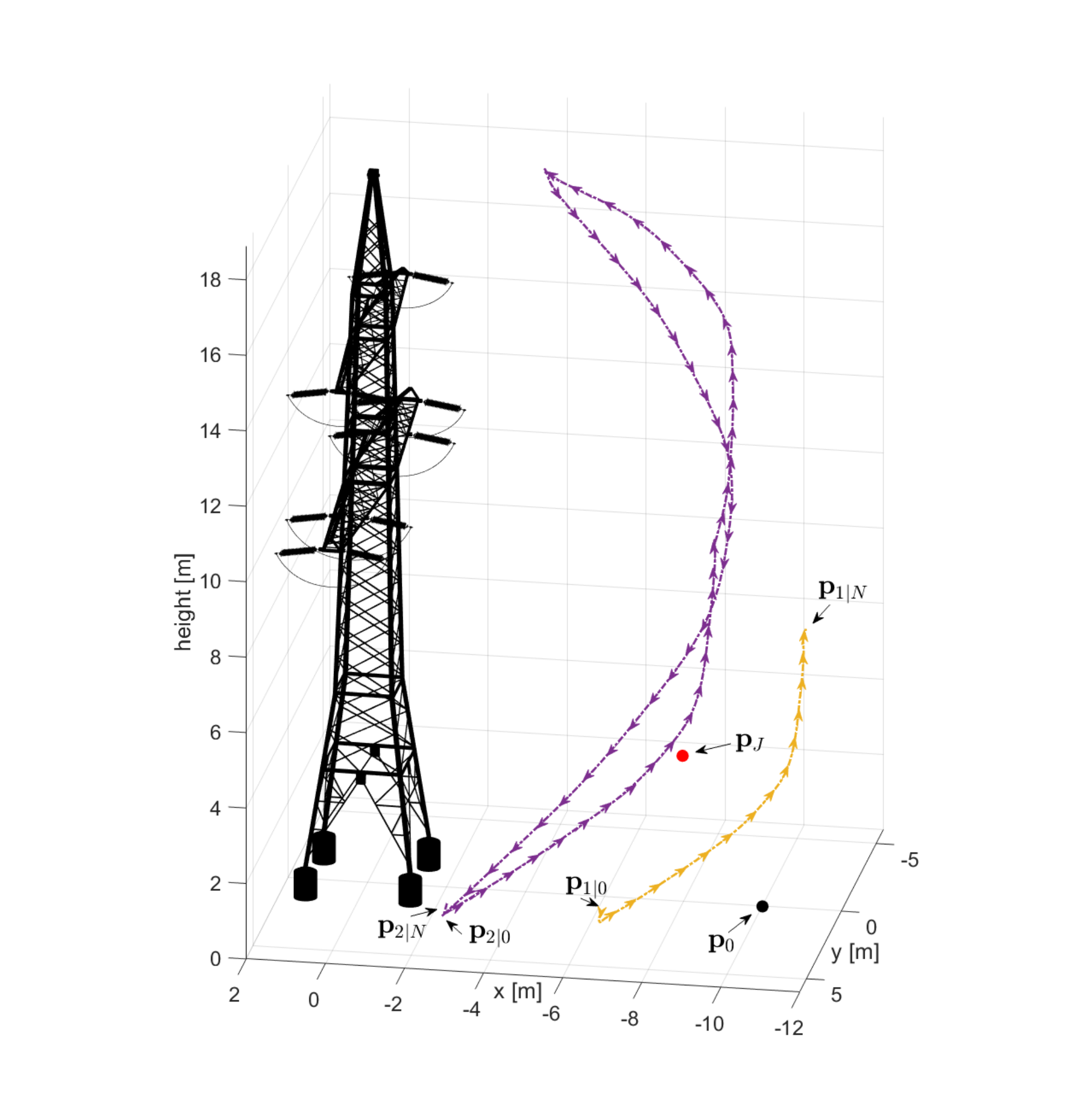}
    \vspace{-0.5em}
    \caption{Simulation overview: relay (\ac{MRAV}-1, yellow), source (\ac{MRAV}-2, purple), ground \ac{BS} (black dot), and jammer (red dot). Arrows indicate motion from initial ($\mathbf{p}_{\bullet|0}$) to final ($\mathbf{p}_{\bullet|N}$) positions.}
    \label{fig:3DtrajectoryFullyActuated}
    \vspace*{-1.35em}
\end{figure}

The \acs{NMPC} jointly minimizes tracking error~\eqref{eq:errorReferenceTrajectory}, misalignment~\eqref{eq:misalignmentCost}, and actuator rate variation~\eqref{eq:outputVariationCost}. Figure~\ref{fig:3DtrajectoryFullyActuated} shows a representative
simulation outcome with \ac{MRAV}-1 and \ac{MRAV}-2 trajectories, the \ac{BS}, and the jammer.

\begin{figure}[tb]
    \centering
    \input{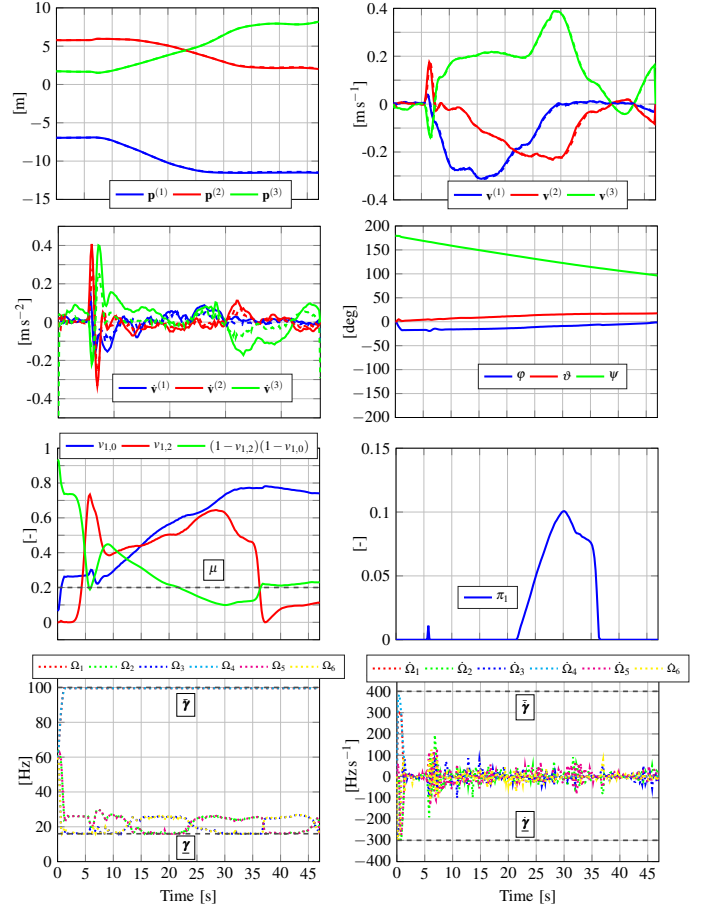}
    \vspace{-0.5em}
    \caption{Time evolution of constrained variables in the \acs{NMPC} problem. Solid lines: references; dashed: measured values. Superscripts $\bullet^{(1)}$, $\bullet^{(2)}$, and $\bullet^{(3)}$ denote $x$-, $y$-, and $z$-axis components.}
    \label{fig:constraints}
    \vspace*{-1.35em}
\end{figure}

Figure~\ref{fig:constraints} presents key state and control variables subject to \acs{NMPC} constraints. Rotor velocities $\bm{\Omega}$ and rates $\dot{\bm{\Omega}}$ remain within $(\underline{\gamma}, \bar{\gamma}, \dot{\underline{\gamma}}, \dot{\bar{\gamma}})$, confirming actuation compliance. A brief activation of slack $\pi$ corresponds to a minor relaxation of~\eqref{eq:nmpc_alignment_constraint} under transient geometry. The mutual misalignment metric remains above $\mu$ for most of
the mission, indicating that antenna-induced channel degradation is contained despite aggressive maneuvers.

From Figure~\ref{fig:constraints}, the translational velocity of \ac{MRAV}-1 remains within $\pm\SI{0.4}{\meter\per\second}$ per axis, giving $\bar{v} = \sqrt{3\times0.4^2} \approx \SI{0.69}{\meter\per\second}$. With $T_s = \SI{15}{\milli\second}$, the per-step displacement is ${\approx}\SI{1.04}{\centi\meter}$, so $\|\bm{\kappa}-\bm{\kappa}_e\|^3 \sim \pazocal{O}(10^{-6})\,\si{\meter^3}$, heavily suppressing the error in~\eqref{eq:taylorRemainder}. The empirical maximum relative error
$|\bar{f}-f^{[2]}|/|\bar{f}|$ was $0.015\%$ over the mission, against a worst-case bound of $0.85\%$ over $\pazocal{D}$. 

\begin{figure}[tb]
    \centering
\hspace{-0.225cm}
\begin{subfigure}{0.425\columnwidth}
    \hspace*{-0.795cm}
    \centering
    \scalebox{0.62}{
    \begin{tikzpicture}
    \begin{axis}[%
    width=2.2119in,%
    height=1.6183in,%
    at={(0.758in,0.481in)},%
    scale only axis,%
    xmin=0,%
    xmax=47.0,%
    ymax=0.1,%
    ymin=-0.1,%
    xmajorgrids,%
    ymajorgrids,%
    ylabel style={yshift=-0.455cm, xshift=-0cm}, 
    ylabel={[\si{\meter}]},%
    ytick={-.1,-.05,0,.05,.1},%
    yticklabels={-0.1,-0.05,0,0.05,0.1},%
    xtick={0,5,...,45},%
    xticklabels={ , , },%
    axis background/.style={fill=white},%
    legend style={at={(0.55,0.20)},anchor=north,legend cell align=left,draw=none,legend 
    columns=-1,align=left,draw=white!15!black}
    ]
    \addplot [color=blue, solid, line width=1.25pt] 
        file{matlabPlots/tracking_errors/e_pos_x.txt};
    \addplot [color=red, solid, line width=1.25pt] 
        file{matlabPlots/tracking_errors/e_pos_y.txt};
    \addplot [color=green, solid, line width=1.25pt] 
        file{matlabPlots/tracking_errors/e_pos_z.txt};
    %
    %
    \legend{\footnotesize{$\mathbf{e}_{p^{(1)}}$}, \footnotesize{$\mathbf{e}_{p^{(2)}}$}, 
        \footnotesize{$\mathbf{e}_{p^{(3)}}$}};%
    \end{axis}
    \end{tikzpicture}
    }
\end{subfigure}
\hspace*{-0.30cm}
\begin{subfigure}{0.425\columnwidth}
    \hspace*{0.095cm}
    \vspace*{0.15cm}
    \centering
    \scalebox{0.62}{
    \begin{tikzpicture}
    \begin{axis}[%
    width=2.2119in,%
    height=1.6183in,%
    at={(0.758in,0.481in)},%
    scale only axis,%
    xmin=0,%
    xmax=47.0,%
    ymax=0.075,%
    ymin=-0.075,%
    xmajorgrids,%
    ymajorgrids,%
    ylabel style={yshift=-0.555cm, xshift=0cm}, 
    ylabel={[\si{\meter\per\second}]},%
    ytick={-.1,-.05,0,.05,.1},%
    yticklabels={-0.1,-0.05,0,0.05,0.1},%
    xtick={0,5,...,45},%
    xticklabels={ , , },%
    axis background/.style={fill=white},%
    legend style={at={(0.55,0.20)},anchor=north,legend cell align=left,draw=none,legend 
    columns=-1,align=left,draw=white!15!black}
    ]
    \addplot [color=blue, solid, line width=1.25pt] 
        file{matlabPlots/tracking_errors/e_vel_x.txt};%
    \addplot [color=red, solid, line width=1.25pt] 
        file{matlabPlots/tracking_errors/e_vel_y.txt};%
    \addplot [color=green, solid, line width=1.25pt] 
        file{matlabPlots/tracking_errors/e_vel_z.txt};%
    %
    %
    \legend{\footnotesize{$\mathbf{e}_{v^{(1)}}$}, \footnotesize{$\mathbf{e}_{v^{(2)}}$}, 
        \footnotesize{$\mathbf{e}_{v^{(3)}}$}};%
    \end{axis}
    \end{tikzpicture}
    }
\end{subfigure}
\\
\vspace{0.05cm}
\hspace{-0.15cm}
\begin{subfigure}{0.425\columnwidth}
    \hspace*{-0.73cm}
    \centering
    \scalebox{0.62}{
    \begin{tikzpicture}
    \begin{axis}[%
    width=2.2119in,%
    height=1.6183in,%
    at={(0.758in,0.481in)},%
    scale only axis,%
    xmin=0,%
    xmax=47.0,%
    ymax=0.25,%
    ymin=-0.25,%
    xmajorgrids,%
    ymajorgrids,%
    ylabel style={yshift=-0.355cm, xshift=0cm}, 
    xlabel={Time [\si{\second}]},%
    ylabel={[-]},%
    ytick={-.5,-.4,-.3,-.2,-.1,0,.1,.2,.3,.4,.5},%
    yticklabels={\empty,-0.4,\empty,-0.2,-0.1,0,0.1,0.2,\empty,0.4,\empty},%
    xtick={0,5,...,45},%
    axis background/.style={fill=white},%
    legend style={at={(0.55,0.20)},anchor=north,legend cell align=left,draw=none,legend 
    columns=-1,align=left,draw=white!15!black}
    ]
    \addplot [color=blue, solid, line width=1.25pt] 
        file{matlabPlots/tracking_errors/e_acc_x.txt};%
    \addplot [color=red, solid, line width=1.25pt] 
        file{matlabPlots/tracking_errors/e_acc_y.txt};%
    \addplot [color=green, solid, line width=1.25pt] 
        file{matlabPlots/tracking_errors/e_acc_z.txt};%
    %
    %
    \legend{\footnotesize{$\mathbf{e}_{\rho^{(1)}}$}, \footnotesize{$\mathbf{e}_{\rho^{(2)}}$}, 
        \footnotesize{$\mathbf{e}_{\rho^{(3)}}$}};%
    \end{axis}
    \end{tikzpicture}
    }
\end{subfigure}
\hspace*{-0.46cm}
\begin{subfigure}{0.425\columnwidth}
    \hspace*{0.16cm}
    \centering
    \scalebox{0.62}{
    \begin{tikzpicture}
    \begin{axis}[%
    width=2.2119in,%
    height=1.6183in,%
    at={(0.758in,0.481in)},%
    scale only axis,%
    xmin=0,%
    xmax=47.0,%
    ymax=1.1,%
    ymin=0,%
    xmajorgrids,%
    ymajorgrids,%
    ylabel style={yshift=-0.055cm, xshift=0cm}, 
    xlabel={Time [\si{\second}]},%
    ylabel={[-]},%
    ytick={0,.2,.4,.6,.8,1},%
    xtick={0,5,...,45},%
    axis background/.style={fill=white},%
    legend style={at={(0.37,0.20)},anchor=north,legend cell align=left,draw=none,legend 
    columns=-1,align=left,draw=white!15!black}
    ]
    \addplot [color=blue, solid, line width=1.25pt] 
        file{matlabPlots/tracking_errors/e_v10.txt};%
    \addplot [color=red, solid, line width=1.25pt] 
        file{matlabPlots/tracking_errors/e_v12.txt};%
    %
    %
    \legend{\footnotesize{$\mathbf{e}_\mathrm{mis}^{(1)}$}, \footnotesize{$\mathbf{e}_\mathrm{mis}^{(2)}$}};%
    \end{axis}
    \end{tikzpicture}
    }
\end{subfigure}
    \vspace{-0.5em}
    \caption{Time evolution of tracking errors in position, velocity, and thrust direction, with antenna misalignment penalty components.}
    \label{fig:trackingErrors}
    \vspace*{-1.35em}
\end{figure}
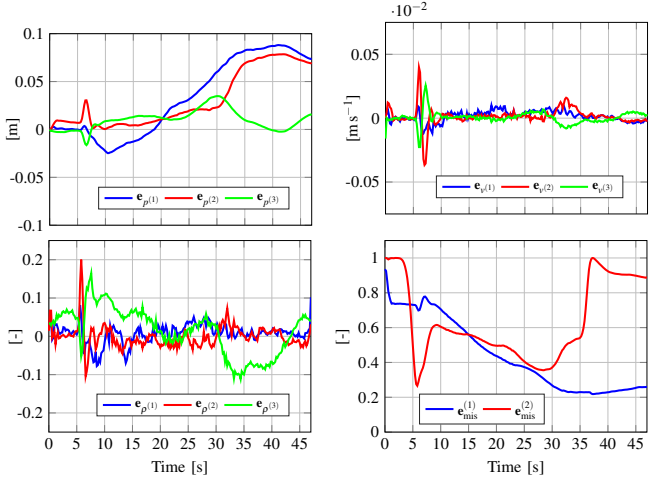

Figure~\ref{fig:trackingErrors} shows tracking errors in position, velocity, and thrust direction, together with misalignment
penalties~\eqref{eq:misalignmentCost}. 
The velocity $\mathbf{e}_v$ and thrust-direction $\mathbf{e}_\rho$ errors remain bounded and converge rapidly after short transients, confirming accurate dynamic regulation. The misalignment penalties $e_\mathrm{mis}^{(1)}$ and $e_\mathrm{mis}^{(2)}$
stay below $0.6$ for most of the mission, with brief peaks during fast reorientations, confirming that communication links are consistently preserved within acceptable alignment margins.

Figure~\ref{fig:twoHopCapacity} shows instantaneous spectral efficiencies $\log_2(1+\Gamma_{1,0})$ and $\log_2(1+\Gamma_{2,1})$ and the resulting end-to-end capacity~\eqref{eq:balancedCapacity} for the constrained and unconstrained formulations; in the unconstrained case, penalties~\eqref{eq:misalignmentCost} and constraint~\eqref{eq:nmpc_alignment_constraint} are removed
from~\eqref{eq:NMPC_formulation}. The main impact is on reliability: the minimum instantaneous end-to-end capacity increases from $7.29\times10^{-5}$ ($C^{(\mathrm{unconstr.})}$) to $6.8\times10^{-3}$ ($C^{(\mathrm{constr.})}$), nearly two orders of magnitude, converting near-outage events into usable operating points.

Average gains are more moderate: $C^{(\mathrm{constr.})}$ exceeds $C^{(\mathrm{unconstr.})}$ by $1.7\%$, while $\log_2(1+\Gamma_{1,0})$ and $\log_2(1+\Gamma_{2,1})$ increase by $16.9\%$ and $5\%$, respectively. This is expected since the optimization targets end-to-end robustness bottlenecked by the weaker link: local trade-offs between links may occur, but deep fades are strongly suppressed and outage probability is reduced---the key requirement for real-time inspection data streaming. These benefits are expected to grow with more directive antennas.

\begin{figure}[tb]
    \centering
\hspace{-1.025cm}
\begin{subfigure}{0.425\columnwidth}
    \hspace*{0.095cm}
    \centering
    \scalebox{0.62}{
    \begin{tikzpicture}
    \begin{axis}[%
    width=2.2119in,%
    height=1.8183in,%
    at={(0.758in,0.481in)},%
    scale only axis,%
    xmin=0,%
    xmax=47.0,%
    ymax=1,%
    ymin=0,%
    xmajorgrids,%
    ymajorgrids,%
    ylabel style={yshift=-0.055cm, xshift=0cm}, 
    xlabel={Time [\si{\second}]},%
    ylabel={Spectral Efficiency [\si{\bit\per\second\per\hertz}]},%
    ytick={0,0.1,0.2,0.3,0.4,0.5,0.6,0.7,0.8,0.9,1},%
    yticklabels={0,\empty,0.2,\empty,0.4,\empty,0.6,\empty,0.8,\empty,1},%
    xtick={0,5,...,45},%
    axis background/.style={fill=white},%
    legend style={at={(0.495,1.17)},anchor=north,legend cell align=left,draw=none,legend 
    columns=-1,align=left,draw=white!15!black}
    ]
    \addplot [color=blue, solid, line width=1.25pt] 
        file{matlabPlots/end-to-end_capacity/C_10_scope.txt};%
    \addplot [color=blue, dashed, line width=1.25pt] 
        file{matlabPlots/end-to-end_capacity/C_10_no_mis_scope.txt};%
    \legend{\footnotesize{$\log_2(1+\Gamma_{1,0})^{(\mathrm{constr.})}$}, \footnotesize{$\log_2(1+\Gamma_{1,0})^{(\mathrm{unconstr.})}$}};%
    \end{axis}
    \end{tikzpicture}
    }
\end{subfigure}
\hspace*{0.60cm}
\begin{subfigure}{0.425\columnwidth}
    \hspace*{0.095cm}
    \centering
    \scalebox{0.62}{
    \begin{tikzpicture}
    \begin{axis}[%
    width=2.2119in,%
    height=1.8183in,%
    at={(0.758in,0.481in)},%
    scale only axis,%
    xmin=0,%
    xmax=47.0,%
    ymax=1,%
    ymin=0,%
    xmajorgrids,%
    ymajorgrids,%
    ylabel style={yshift=-0.055cm, xshift=0cm}, 
    xlabel={Time [\si{\second}]},%
    ylabel={Spectral Efficiency [\si{\bit\per\second\per\hertz}]},%
    ytick={0,0.1,0.2,0.3,0.4,0.5,0.6,0.7,0.8,0.9,1},%
    yticklabels={0,\empty,0.2,\empty,0.4,\empty,0.6,\empty,0.8,\empty,1},%
    xtick={0,5,...,45},%
    axis background/.style={fill=white},%
    legend style={at={(0.495,1.17)},anchor=north,legend cell align=left,draw=none,legend 
    columns=-1,align=left,draw=white!15!black}
    ]
    \addplot [color=blue, solid, line width=1.25pt] 
        file{matlabPlots/end-to-end_capacity/C_21_scope.txt};%
    \addplot [color=blue, dashed, line width=1.25pt] 
        file{matlabPlots/end-to-end_capacity/C_21_no_mis_scope.txt};%
    \legend{\footnotesize{$\log_2(1+\Gamma_{2,1})^{(\mathrm{constr.})}$}, \footnotesize{$\log_2(1+\Gamma_{2,1})^{(\mathrm{unconstr.})}$}};%
    \end{axis}
    \end{tikzpicture}
    }
\end{subfigure}
\\
\vspace{0.05cm}
\hspace{-0.15cm}
\begin{subfigure}{0.425\columnwidth}
    \hspace*{-0.595cm}
    \centering
    \scalebox{0.62}{
    \begin{tikzpicture}
    \begin{axis}[%
    width=2.2119in,%
    height=1.8183in,%
    at={(0.758in,0.481in)},%
    scale only axis,%
    xmin=0,%
    xmax=47.0,%
    ymax=0.5,%
    ymin=0,%
    xmajorgrids,%
    ymajorgrids,%
    ylabel style={yshift=-0.055cm, xshift=-0cm}, 
    xlabel={Time [\si{\second}]},%
    ylabel={Normalized bit rate [-]},%
    ytick={0,0.1,0.2,0.3,0.4,0.5,0.6,0.7,0.8,0.9,1},%
    yticklabels={0,0.1,0.2,0.3,0.4,0.5,0.6,\empty,0.8,\empty,1},%
    xtick={0,5,...,45},%
    axis background/.style={fill=white},%
    legend style={at={(0.60,0.86)},anchor=north,legend cell align=left,draw=none,legend 
    columns=-1,align=left,draw=white!15!black}
    ]
    \addplot [color=blue, solid, line width=1.25pt] 
        file{matlabPlots/end-to-end_capacity/C.txt};%
    \addplot [color=blue, dashed, line width=1.25pt] 
        file{matlabPlots/end-to-end_capacity/C_no_mis.txt};%
    \legend{\footnotesize{$C^{(\mathrm{constr.})}$}, \footnotesize{$C^{(\mathrm{unconstr.})}$}};%
    \end{axis}
    \end{tikzpicture}
    }
\end{subfigure}
    \vspace{-0.5em}
    \caption{Instantaneous spectral efficiencies $\log_2(1+\Gamma_{1,0})$ and $\log_2(1+\Gamma_{2,1})$ and end-to-end channel capacity \eqref{eq:balancedCapacity}, comparing constrained and unconstrained formulations.}
    \label{fig:twoHopCapacity}
    \vspace*{-1.35em}
\end{figure}


\subsection{Coplanar and tilted-propeller \ac{MRAV} configurations}
\label{sec:coplanerVsTiltedConfigurations}

This section examines how coplanar versus tilted-propeller designs affect the controller's ability to maintain directional communication, isolating the role of actuation geometry in preserving antenna alignment under interference. 

In the coplanar case, all propellers lie on a single plane and generate thrust only along the body $\zz_{B}$-axis, so any lateral motion requires platform tilting, inherently coupling translation with antenna orientation. The tilted-propeller configuration provides partial decoupling between translation and rotation, enabling direct lateral thrust with minimal attitude changes and thus more favorable conditions for maintaining alignment.

Both vehicles are modeled using the \ac{GTMR} framework \cite{Ryll2019IJRR, HamandiIJRR2021}, keeping the \acs{NMPC} structure
identical across platforms so that differences in feasible wrench sets isolate the influence of mechanical actuation alone.

\begin{figure}[tb]
    \centering
    \begin{subfigure}{0.52\columnwidth}
        \centering
        \adjincludegraphics[width=\columnwidth, trim={{.075\width} {.070\height} {.070\width} {.075\height}}, clip]{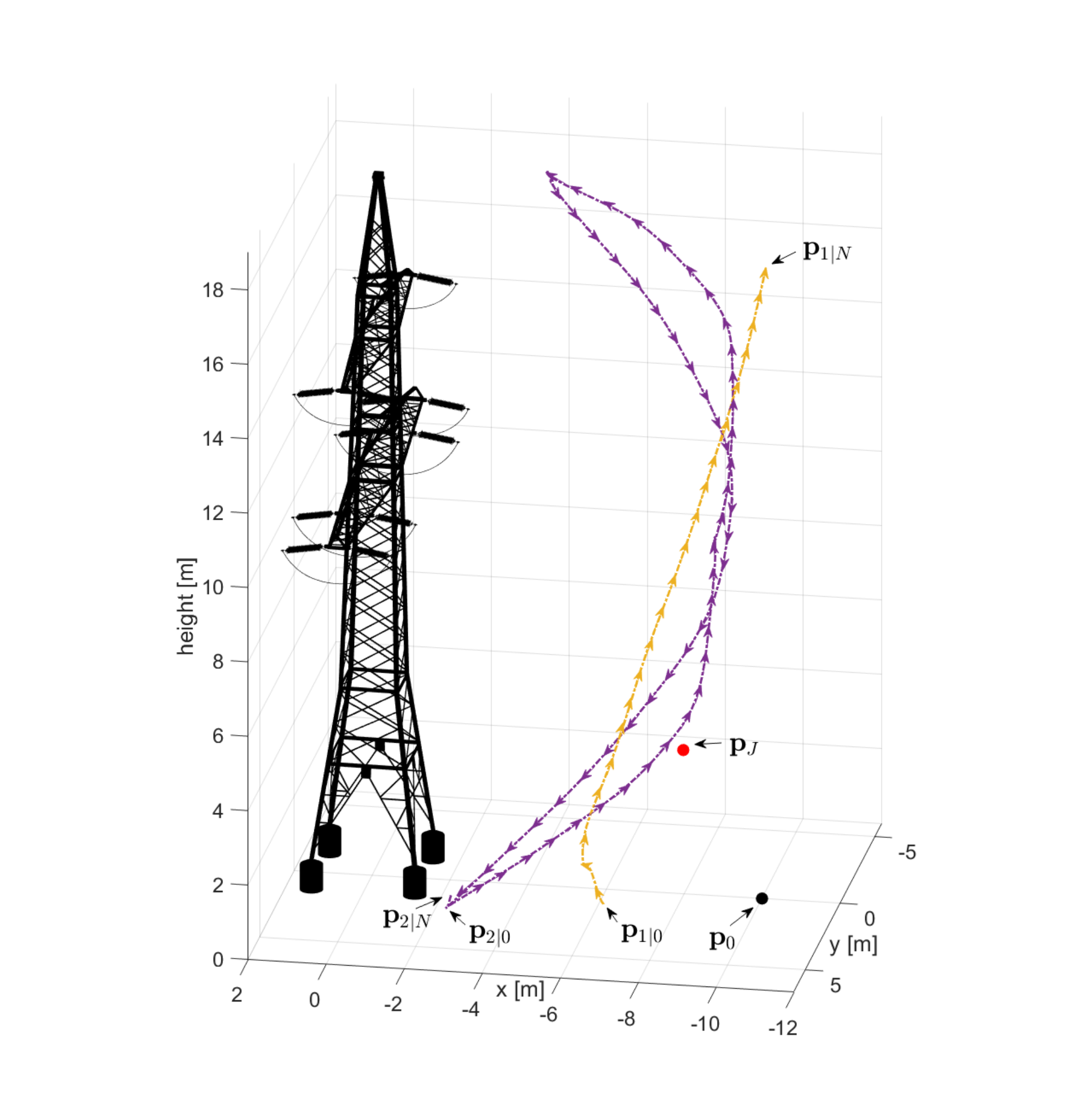}
        \vspace{-1.5em}
        \caption{}
        \label{fig:scenario16Under}
        \end{subfigure}
    \hspace*{-1.75em}
    \begin{subfigure}{0.52\columnwidth}
        \centering 
        \adjincludegraphics[width=\columnwidth, trim={{.075\width} {.070\height} {.070\width} {.075\height}}, clip]{figure/scenario_16_jammer_ON_fully.pdf}
        \vspace{-1.5em}
        \caption{}
        \label{fig:scenario16Fully}
    \end{subfigure}
    \\
    \begin{subfigure}{0.52\columnwidth}
        \centering
        \adjincludegraphics[width=\columnwidth, trim={{.075\width} {.070\height} {.070\width} {.075\height}}, clip]{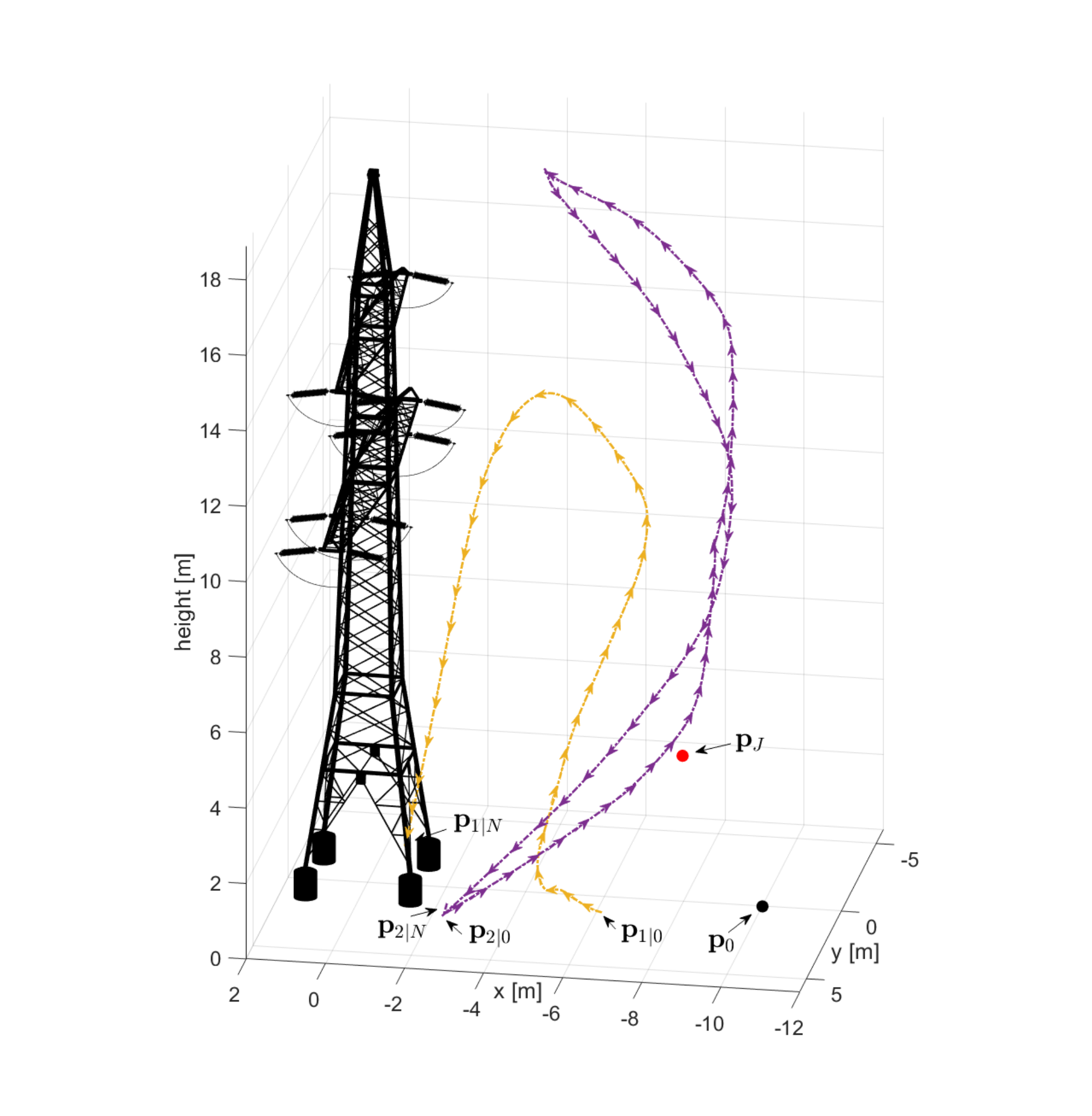}
        \vspace{-1.5em}
        \caption{}
        \label{fig:scenario17Under}
        \end{subfigure}
    \hspace*{-1.75em}
    \begin{subfigure}{0.52\columnwidth}
        \centering
        \adjincludegraphics[width=\columnwidth, trim={{.075\width} {.070\height} {.070\width} {.075\height}}, clip]{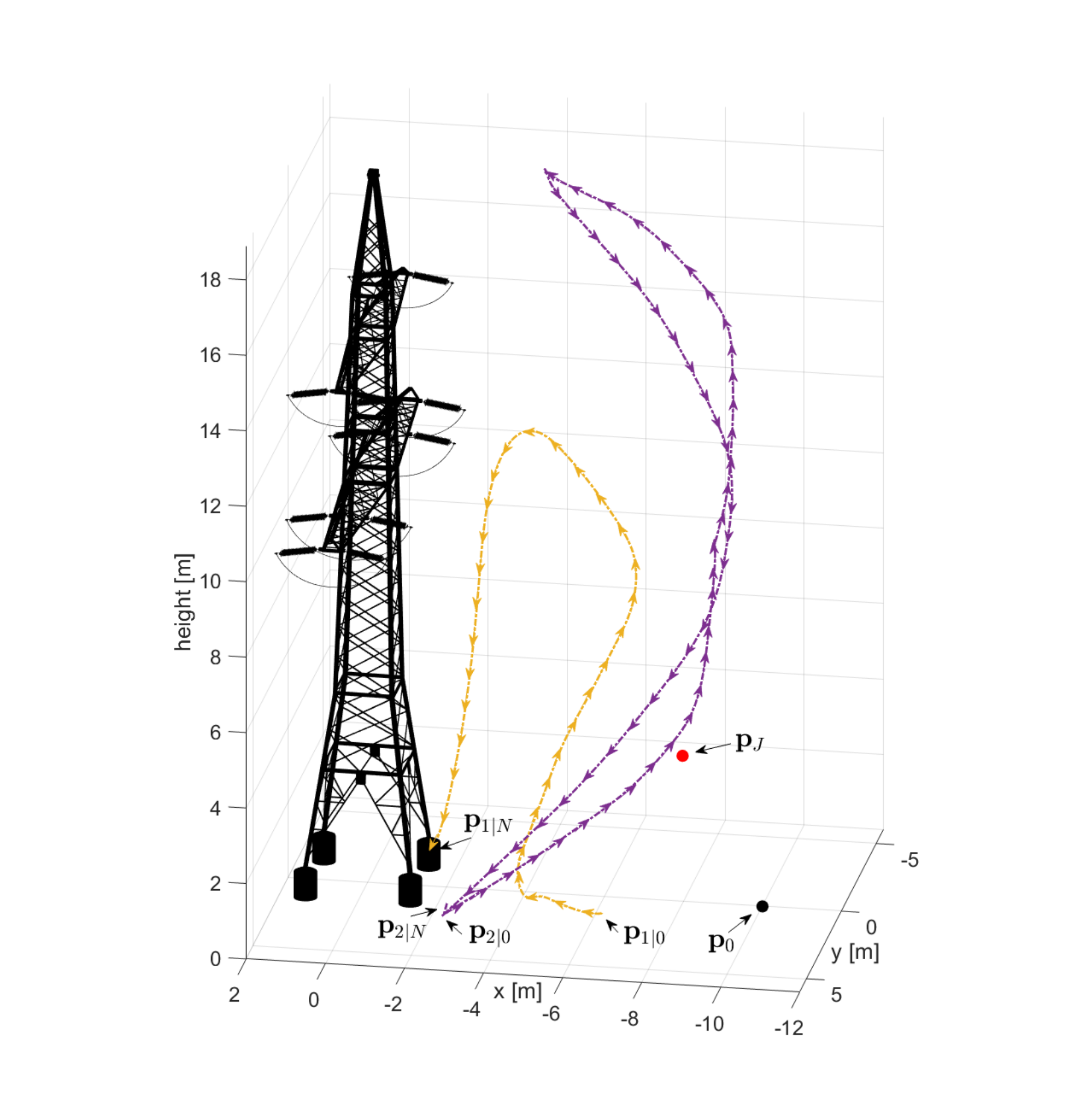}
        \vspace{-1.5em}
        \caption{}
        \label{fig:scenario17Fully}
    \end{subfigure}
    \vspace*{-0.25em}
    \caption{Simulation overviews for coplanar (\ref{fig:scenario16Under},\,\ref{fig:scenario17Under}) and tilted-propeller (\ref{fig:scenario16Fully},\,\ref{fig:scenario17Fully}) configurations, with (\ref{fig:scenario16Under},\,\ref{fig:scenario16Fully}) and without (\ref{fig:scenario17Under},\,\ref{fig:scenario17Fully}) antenna misalignment cost and constraints.} 
    \label{fig:3DtrajectoryJammerAlwaysON}
    \vspace*{-1.35em}
\end{figure}

Figure~\ref{fig:3DtrajectoryJammerAlwaysON} compares constrained and unconstrained scenarios. Without alignment penalties, the relay exhibits large attitude excursions and wide lateral deviations, producing longer, less direct trajectories. With alignment constraints enforced, the relay actively regulates position and orientation to preserve dual link alignment, yielding smoother, more compact paths close to the communication axis.

The constrained trajectories reveal the clearest platform differences. In the coplanar configuration (Figure~\ref{fig:scenario16Under}), lateral motion requires significant body tilting, coupling translation and antenna pointing, reducing simultaneous alignment with both peers, and driving the vehicle toward actuation limits. To maintain acceptable alignment, the \acs{NMPC} increases altitude to about \SI{16}{\meter}, where \ac{LoS} geometry enlarges the feasible intersection of the two alignment cones. By contrast, the tilted-propeller configuration (Figure~\ref{fig:scenario16Fully}) generates lateral forces with minimal attitude changes, preserving beam orientation and allowing a shorter, lower trajectory (ending near \SI{8}{\meter}) while fully satisfying alignment and actuation constraints.

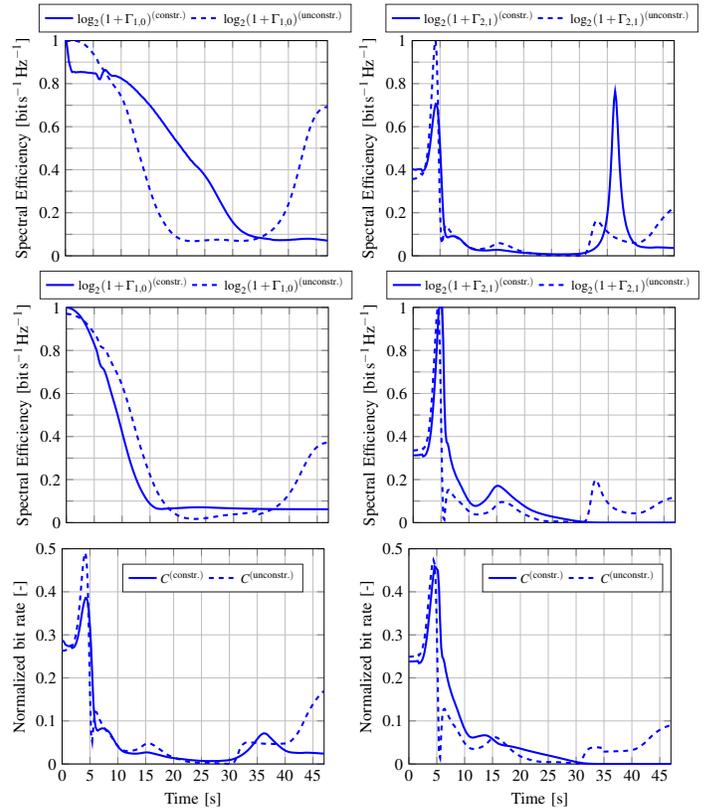
\begin{figure}[tb]
    \centering
\hspace{-1.025cm}
\begin{subfigure}{0.425\columnwidth}
    \hspace*{0.095cm}
    \centering
    \scalebox{0.62}{
    \begin{tikzpicture}
    \begin{axis}[%
    width=2.2119in,%
    height=1.8183in,%
    at={(0.758in,0.481in)},%
    scale only axis,%
    xmin=0,%
    xmax=47.0,%
    ymax=1,%
    ymin=0,%
    xmajorgrids,%
    ymajorgrids,%
    ylabel style={yshift=-0.055cm, xshift=0cm}, 
    extra x tick labels={ , ,  , },%
    ylabel={Spectral Efficiency [\si{\bit\per\second\per\hertz}]},%
    ytick={0,0.1,0.2,0.3,0.4,0.5,0.6,0.7,0.8,0.9,1},%
    yticklabels={0,\empty,0.2,\empty,0.4,\empty,0.6,\empty,0.8,\empty,1},%
    xtick={0,5,...,45},%
    xticklabels={ , , },%
    axis background/.style={fill=white},%
    legend style={at={(0.495,1.17)},anchor=north,legend cell align=left,draw=none,legend 
    columns=-1,align=left,draw=white!15!black}
    ]
    \addplot [color=blue, solid, line width=1.25pt] 
        file{matlabPlots/end-to-end_capacity/C_10_scope.txt};%
    \addplot [color=blue, dashed, line width=1.25pt] 
        file{matlabPlots/end-to-end_capacity/C_10_no_mis_scope.txt};%
    \legend{\footnotesize{$\log_2(1+\Gamma_{1,0})^{(\mathrm{constr.})}$}, \footnotesize{$\log_2(1+\Gamma_{1,0})^{(\mathrm{unconstr.})}$}};%
    \end{axis}
    \end{tikzpicture}
    }
\end{subfigure}
\hspace*{0.60cm}
\begin{subfigure}{0.425\columnwidth}
    \hspace*{0.095cm}
    \centering
    \scalebox{0.62}{
    \begin{tikzpicture}
    \begin{axis}[%
    width=2.2119in,%
    height=1.8183in,%
    at={(0.758in,0.481in)},%
    scale only axis,%
    xmin=0,%
    xmax=47.0,%
    ymax=1,%
    ymin=0,%
    xmajorgrids,%
    ymajorgrids,%
    ylabel style={yshift=-0.055cm, xshift=0cm}, 
    extra x tick labels={ , ,  , },%
    ylabel={Spectral Efficiency [\si{\bit\per\second\per\hertz}]},%
    ytick={0,0.1,0.2,0.3,0.4,0.5,0.6,0.7,0.8,0.9,1},%
    yticklabels={0,\empty,0.2,\empty,0.4,\empty,0.6,\empty,0.8,\empty,1},%
    xtick={0,5,...,45},%
    xticklabels={ , , },%
    axis background/.style={fill=white},%
    legend style={at={(0.495,1.17)},anchor=north,legend cell align=left,draw=none,legend 
    columns=-1,align=left,draw=white!15!black}
    ]
    \addplot [color=blue, solid, line width=1.25pt] 
        file{matlabPlots/end-to-end_capacity/C_21_scope.txt};%
    \addplot [color=blue, dashed, line width=1.25pt] 
        file{matlabPlots/end-to-end_capacity/C_21_no_mis_scope.txt};%
    \legend{\footnotesize{$\log_2(1+\Gamma_{2,1})^{(\mathrm{constr.})}$}, \footnotesize{$\log_2(1+\Gamma_{2,1})^{(\mathrm{unconstr.})}$}};%
    \end{axis}
    \end{tikzpicture}
    }
\end{subfigure}
\\
\vspace{0.05cm}
\hspace{-1.0cm}
\begin{subfigure}{0.425\columnwidth}
    \hspace*{0.095cm}
    \centering
    \scalebox{0.62}{
    \begin{tikzpicture}
    \begin{axis}[%
    width=2.2119in,%
    height=1.8183in,%
    at={(0.758in,0.481in)},%
    scale only axis,%
    xmin=0,%
    xmax=47.0,%
    ymax=1,%
    ymin=0,%
    xmajorgrids,%
    ymajorgrids,%
    ylabel style={yshift=-0.055cm, xshift=0cm}, 
    ylabel={Spectral Efficiency [\si{\bit\per\second\per\hertz}]},%
    ytick={0,0.1,0.2,0.3,0.4,0.5,0.6,0.7,0.8,0.9,1},%
    yticklabels={0,\empty,0.2,\empty,0.4,\empty,0.6,\empty,0.8,\empty,1},%
    xtick={0,5,...,45},%
    xticklabels={ , , },%
    axis background/.style={fill=white},%
    legend style={at={(0.495,1.17)},anchor=north,legend cell align=left,draw=none,legend 
    columns=-1,align=left,draw=white!15!black}
    ]
    \addplot [color=blue, solid, line width=1.25pt] 
        file{matlabPlots/end-to-end_capacity/C_10_scope_under.txt};%
    \addplot [color=blue, dashed, line width=1.25pt] 
        file{matlabPlots/end-to-end_capacity/C_10_no_mis_scope_under.txt};%
    \legend{\footnotesize{$\log_2(1+\Gamma_{1,0})^{(\mathrm{constr.})}$}, \footnotesize{$\log_2(1+\Gamma_{1,0})^{(\mathrm{unconstr.})}$}};%
    \end{axis}
    \end{tikzpicture}
    }
\end{subfigure}
\hspace*{0.60cm}
\begin{subfigure}{0.425\columnwidth}
    \hspace*{0.095cm}
    \centering
    \scalebox{0.62}{
    \begin{tikzpicture}
    \begin{axis}[%
    width=2.2119in,%
    height=1.8183in,%
    at={(0.758in,0.481in)},%
    scale only axis,%
    xmin=0,%
    xmax=47.0,%
    ymax=1,%
    ymin=0,%
    xmajorgrids,%
    ymajorgrids,%
    ylabel style={yshift=-0.055cm, xshift=0cm}, 
    ylabel={Spectral Efficiency [\si{\bit\per\second\per\hertz}]},%
    ytick={0,0.1,0.2,0.3,0.4,0.5,0.6,0.7,0.8,0.9,1},%
    yticklabels={0,\empty,0.2,\empty,0.4,\empty,0.6,\empty,0.8,\empty,1},%
    xtick={0,5,...,45},%
    xticklabels={ , , },%
    axis background/.style={fill=white},%
    legend style={at={(0.495,1.17)},anchor=north,legend cell align=left,draw=none,legend 
    columns=-1,align=left,draw=white!15!black}
    ]
    \addplot [color=blue, solid, line width=1.25pt] 
        file{matlabPlots/end-to-end_capacity/C_21_scope_under.txt};%
    \addplot [color=blue, dashed, line width=1.25pt] 
        file{matlabPlots/end-to-end_capacity/C_21_no_mis_scope_under.txt};%
    \legend{\footnotesize{$\log_2(1+\Gamma_{2,1})^{(\mathrm{constr.})}$}, \footnotesize{$\log_2(1+\Gamma_{2,1})^{(\mathrm{unconstr.})}$}};%
    \end{axis}
    \end{tikzpicture}
    }
\end{subfigure}
\\
\vspace{0.05cm}
\hspace{-0.755cm}
\begin{subfigure}{0.425\columnwidth}
    \hspace*{0.095cm}
    \centering
    \scalebox{0.62}{
    \begin{tikzpicture}
    \begin{axis}[%
    width=2.2119in,%
    height=1.8183in,%
    at={(0.758in,0.481in)},%
    scale only axis,%
    xmin=0,%
    xmax=47.0,%
    ymax=0.5,%
    ymin=0,%
    xmajorgrids,%
    ymajorgrids,%
    ylabel style={yshift=-0.055cm, xshift=0cm}, 
    xlabel={Time [\si{\second}]},%
    ylabel={Normalized bit rate [-]},%
    ytick={0,0.1,0.2,0.3,0.4,0.5,0.6,0.7,0.8,0.9,1},%
    yticklabels={0,0.1,0.2,0.3,0.4,0.5,0.6,\empty,0.8,\empty,1},%
    xtick={0,5,...,45},%
    axis background/.style={fill=white},%
    legend style={at={(0.575,0.93)},anchor=north,legend cell align=left,draw=none,legend 
    columns=-1,align=left,draw=white!15!black}
    ]
    \addplot [color=blue, solid, line width=1.25pt] 
        file{matlabPlots/end-to-end_capacity/C.txt};%
    \addplot [color=blue, dashed, line width=1.25pt] 
        file{matlabPlots/end-to-end_capacity/C_no_mis.txt};%
    \legend{\footnotesize{$C^{(\mathrm{constr.})}$}, \footnotesize{$C^{(\mathrm{unconstr.})}$}};%
    \end{axis}
    \end{tikzpicture}
    }
\end{subfigure}
\hspace*{0.60cm}
\begin{subfigure}{0.425\columnwidth}
    \hspace*{0.095cm}
    \centering
    \scalebox{0.62}{
    \begin{tikzpicture}
    \begin{axis}[%
    width=2.2119in,%
    height=1.8183in,%
    at={(0.758in,0.481in)},%
    scale only axis,%
    xmin=0,%
    xmax=47.0,%
    ymax=0.5,%
    ymin=0,%
    xmajorgrids,%
    ymajorgrids,%
    ylabel style={yshift=-0.055cm, xshift=0cm}, 
    xlabel={Time [\si{\second}]},%
    ylabel={Normalized bit rate [-]},%
    ytick={0,0.1,0.2,0.3,0.4,0.5,0.6,0.7,0.8,0.9,1},%
    yticklabels={0,0.1,0.2,0.3,0.4,0.5,0.6,\empty,0.8,\empty,1},%
    xtick={0,5,...,45},%
    axis background/.style={fill=white},%
    legend style={at={(0.625,0.93)},anchor=north,legend cell align=left,draw=none,legend 
    columns=-1,align=left,draw=white!15!black}
    ]
    \addplot [color=blue, solid, line width=1.25pt] 
        file{matlabPlots/end-to-end_capacity/C_under.txt};%
    \addplot [color=blue, dashed, line width=1.25pt] 
        file{matlabPlots/end-to-end_capacity/C_no_mis_under.txt};%
    \legend{\footnotesize{$C^{(\mathrm{constr.})}$}, \footnotesize{$C^{(\mathrm{unconstr.})}$}};%
    \end{axis}
    \end{tikzpicture}
    }
\end{subfigure}
    \vspace{-0.5em}
    \caption{Instantaneous spectral efficiencies and end-to-end capacity for constrained and unconstrained formulations. Top row: tilted-propeller; middle row: coplanar. Bottom row: end-to-end capacities side-by-side for tilted-propeller (left) and coplanar (right).} 
    \label{fig:twoHopCapacityComplete}
    \vspace*{-1.35em}
\end{figure}

Figure~\ref{fig:twoHopCapacityComplete} confirms that actuation geometry strongly affects the communication gains of the constrained formulation. Average end-to-end capacity increases by $1.7\%$ for the tilted platform and by $0.55\%$ for the coplanar platform; the gap reflects the stronger translation---attitude coupling in coplanar designs, which makes simultaneous dual-link optimization more difficult. Quantitatively, the constrained formulation improves relay--\ac{BS} and source--relay spectral
efficiencies by $16.9\%$ and $5\%$ in the tilted case versus $12.49\%$ and $2.41\%$ in the coplanar case, consistent with fewer slack activations and larger feasibility margins in the tilted configuration.

Overall, both platforms remain dynamically feasible, but their communication performance differs substantially: the coplanar configuration experiences progressive translation--attitude coupling that collapses end-to-end capacity after approximately $\SI{30}{\second}$, whereas the tilted-propeller platform preserves alignment throughout, avoiding outages under persistent jamming. Full actuation is therefore a key enabler of reliable directional communication under interference, while coplanar
designs are mechanically ill-suited for sustained operation.

Although energy efficiency was not the focus of this formulation, we briefly quantify the energy trade-off between the two platforms: tilted-propeller designs trade higher steady-state thrust for the actuation redundancy enabling alignment preservation~\cite{Ryll2019IJRR, HamandiIJRR2021}. The total aerodynamic energy $\int_0^{T_{\mathrm{mission}}} \sum_{i=1}^{N_p} c_{\tau_i} \Omega_i^{3/2}\,dt$ over the nominal mission is $\SI{14.6}{\kilo\joule}$ for the tilted-propeller platform ($\alpha=20^\circ$), a $9.8\%$ increase over the $\SI{13.29}{\kilo\joule}$ consumed by the coplanar configuration. 



\subsection{Robustness under time-varying jamming}
\label{sec:robustnessJammingConditions}

This section assesses robustness under time-varying interference. The jammer alternates between active and inactive phases on a periodic ON--OFF schedule ($\SI{5}{\second}$ active, $\SI{5}{\second}$ inactive), with the OFF phase implemented by setting $P_J=0$. This timescale significantly exceeds the vehicle dynamics, enabling meaningful evaluation of the controller's
response to abrupt interference changes.

\begin{figure}[tb]
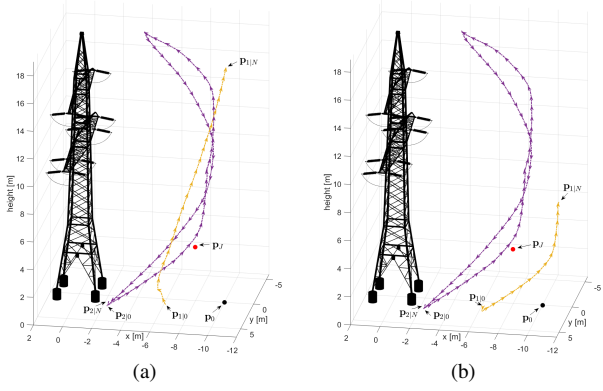

    \centering
    \begin{subfigure}{0.52\columnwidth}
        \centering
        \adjincludegraphics[width=\columnwidth, trim={{.075\width} {.070\height} {.070\width} {.075\height}}, clip]{figure/scenario_16_jammer_ON_under.pdf}
        \vspace{-1.5em}
        \caption{}
        \label{fig:scenario16Under-ONOFF}
        \end{subfigure}
    \hspace*{-1.75em}
    \begin{subfigure}{0.52\columnwidth}
        \centering 
        \adjincludegraphics[width=\columnwidth, trim={{.075\width} {.070\height} {.070\width} {.075\height}}, clip]{figure/scenario_16_jammer_ON_fully.pdf}
        \vspace{-1.5em}
        \caption{}
        \label{fig:scenario16Fully-ONOFF}
    \end{subfigure}
    \vspace*{-0.25em}
    \caption{Simulation overviews for coplanar (\ref{fig:scenario16Under-ONOFF}) and tilted-propeller (\ref{fig:scenario16Fully-ONOFF}) configurations under an ON–OFF jamming.}
    \label{fig:3DtrajectoryJammerONOFF}
    \vspace*{-1.35em}
\end{figure}

The 3D trajectories under ON--OFF jamming (Figure~\ref{fig:3DtrajectoryJammerONOFF}) closely match those from the always-active case, revealing how rapidly the system recovers once the disturbance subsides: the controller maintains nearly the same relative
geometry, altitude, and attitude throughout, choosing a stable, alignment-optimal configuration rather than reacting to transient channel improvements.

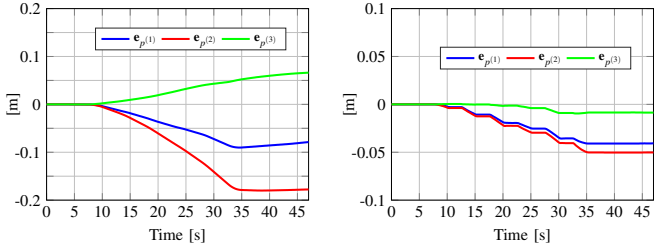
\begin{figure}[tb]
    \centering
\hspace{-0.225cm}
\begin{subfigure}{0.425\columnwidth}
    \hspace*{-0.795cm}
    \centering
    \scalebox{0.62}{
    \begin{tikzpicture}
    \begin{axis}[%
    width=2.2119in,%
    height=1.6183in,%
    at={(0.758in,0.481in)},%
    scale only axis,%
    xmin=0,%
    xmax=47.0,%
    ymax=0.20,%
    ymin=-0.20,%
    xmajorgrids,%
    ymajorgrids,%
    ylabel style={yshift=-0.355cm, xshift=-0cm}, 
    xlabel={Time [\si{\second}]},%
    ylabel={[\si{\meter}]},%
    ytick={-.2,-.15,-.1,-.05,0,.05,.1,.15,.2},%
    yticklabels={-0.2,\empty,-0.1,\empty,0,\empty,0.1,\empty,0.2},%
    xtick={0,5,...,45},%
    axis background/.style={fill=white},%
    legend style={at={(0.55,0.90)},anchor=north,legend cell align=left,draw=none,legend 
    columns=-1,align=left,draw=white!15!black}
    ]
    \addplot [color=blue, solid, line width=1.25pt] 
        file{matlabPlots/tracking_errors_ON-OFF/diff_x_under.txt};
    \addplot [color=red, solid, line width=1.25pt] 
        file{matlabPlots/tracking_errors_ON-OFF/diff_y_under.txt};
    \addplot [color=green, solid, line width=1.25pt] 
        file{matlabPlots/tracking_errors_ON-OFF/diff_z_under.txt};
    %
    %
    \legend{\footnotesize{$\mathbf{e}_{p^{(1)}}$}, \footnotesize{$\mathbf{e}_{p^{(2)}}$}, 
        \footnotesize{$\mathbf{e}_{p^{(3)}}$}};%
    \end{axis}
    \end{tikzpicture}
    }
\end{subfigure}
\hspace*{-0.30cm}
\begin{subfigure}{0.425\columnwidth}
    \hspace*{0.095cm}
    \centering
    \scalebox{0.62}{
    \begin{tikzpicture}
    \begin{axis}[%
    width=2.2119in,%
    height=1.6183in,%
    at={(0.758in,0.481in)},%
    scale only axis,%
    xmin=0,%
    xmax=47.0,%
    ymax=0.1,%
    ymin=-0.1,%
    xmajorgrids,%
    ymajorgrids,%
    ylabel style={yshift=-0.355cm, xshift=0cm}, 
    xlabel={Time [\si{\second}]},%
    ylabel={[\si{\meter}]},%
    ytick={-.1,-.05,0,.05,.1},%
    yticklabels={-0.1,-0.05,0,0.05,0.1},%
    xtick={0,5,...,45},%
    axis background/.style={fill=white},%
    legend style={at={(0.55,0.80)},anchor=north,legend cell align=left,draw=none,legend 
    columns=-1,align=left,draw=white!15!black}
    ]
    \addplot [color=blue, solid, line width=1.25pt] 
        file{matlabPlots/tracking_errors_ON-OFF/diff_x_fully.txt};%
    \addplot [color=red, solid, line width=1.25pt] 
        file{matlabPlots/tracking_errors_ON-OFF/diff_y_fully.txt};%
    \addplot [color=green, solid, line width=1.25pt] 
        file{matlabPlots/tracking_errors_ON-OFF/diff_z_fully.txt};%
    %
    %
    \legend{\footnotesize{$\mathbf{e}_{p^{(1)}}$}, \footnotesize{$\mathbf{e}_{p^{(2)}}$}, 
        \footnotesize{$\mathbf{e}_{p^{(3)}}$}};%
    \end{axis}
    \end{tikzpicture}
    }
\end{subfigure}
    \vspace{-0.5em}
    \caption{Trajectory deviation between the always-ON and ON–OFF jamming scenarios for coplanar (left) and tilted-propeller (right) configurations.}
    \label{fig:trackingErrorsONOFF}
    \vspace*{-1.15em}
\end{figure}

Figure~\ref{fig:trackingErrorsONOFF} quantifies the deviation between always-ON and ON--OFF trajectories, which remains within
$[-0.2,\,0.2]\,\si{\meter}$ for the coplanar platform and $[-0.05,\,0.05]\,\si{\meter}$ for the tilted-propeller one. The small
deviation is consistent with the ON--OFF cycle ($\SI{10}{\second}$) being much longer than the system's dynamic response time, so the optimizer converges to a near-invariant alignment-preserving motion. The smaller deviation for the tilted platform reflects its enhanced actuation redundancy, enabling tighter tracking and less sensitivity to interference transients.

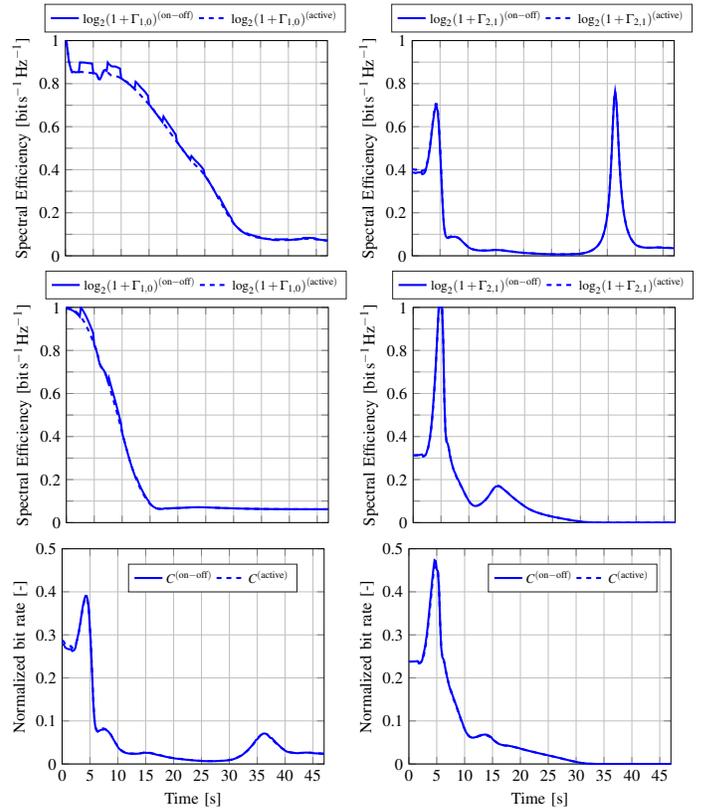
\begin{figure}[tb]
    \centering
\hspace{-1.025cm}
\begin{subfigure}{0.425\columnwidth}
    \hspace*{0.095cm}
    \centering
    \scalebox{0.62}{
    \begin{tikzpicture}
    \begin{axis}[%
    width=2.2119in,%
    height=1.8183in,%
    at={(0.758in,0.481in)},%
    scale only axis,%
    xmin=0,%
    xmax=47.0,%
    ymax=1,%
    ymin=0,%
    xmajorgrids,%
    ymajorgrids,%
    ylabel style={yshift=-0.055cm, xshift=0cm}, 
    extra x tick labels={ , ,  , },%
    ylabel={Spectral Efficiency [\si{\bit\per\second\per\hertz}]},%
    ytick={0,0.1,0.2,0.3,0.4,0.5,0.6,0.7,0.8,0.9,1},%
    yticklabels={0,\empty,0.2,\empty,0.4,\empty,0.6,\empty,0.8,\empty,1},%
    xtick={0,5,...,45},%
    xticklabels={ , , },%
    axis background/.style={fill=white},%
    legend style={at={(0.495,1.17)},anchor=north,legend cell align=left,draw=none,legend 
    columns=-1,align=left,draw=white!15!black}
    ]
    \addplot [color=blue, solid, line width=1.25pt] 
        file{matlabPlots/end-to-end_capacity/C_10_scope_on_off.txt};%
    \addplot [color=blue, dashed, line width=1.25pt] 
        file{matlabPlots/end-to-end_capacity/C_10_scope.txt};%
    \legend{\footnotesize{$\log_2(1+\Gamma_{1,0})^{(\mathrm{on-off})}$}, \footnotesize{$\log_2(1+\Gamma_{1,0})^{(\mathrm{active})}$}};%
    \end{axis}
    \end{tikzpicture}
    }
\end{subfigure}
\hspace*{0.60cm}
\begin{subfigure}{0.425\columnwidth}
    \hspace*{0.095cm}
    \centering
    \scalebox{0.62}{
    \begin{tikzpicture}
    \begin{axis}[%
    width=2.2119in,%
    height=1.8183in,%
    at={(0.758in,0.481in)},%
    scale only axis,%
    xmin=0,%
    xmax=47.0,%
    ymax=1,%
    ymin=0,%
    xmajorgrids,%
    ymajorgrids,%
    ylabel style={yshift=-0.055cm, xshift=0cm}, 
    extra x tick labels={ , ,  , },%
    ylabel={Spectral Efficiency [\si{\bit\per\second\per\hertz}]},%
    ytick={0,0.1,0.2,0.3,0.4,0.5,0.6,0.7,0.8,0.9,1},%
    yticklabels={0,\empty,0.2,\empty,0.4,\empty,0.6,\empty,0.8,\empty,1},%
    xtick={0,5,...,45},%
    xticklabels={ , , },%
    axis background/.style={fill=white},%
    legend style={at={(0.495,1.17)},anchor=north,legend cell align=left,draw=none,legend 
    columns=-1,align=left,draw=white!15!black}
    ]
    \addplot [color=blue, solid, line width=1.25pt] 
        file{matlabPlots/end-to-end_capacity/C_21_scope_on_off.txt};%
    \addplot [color=blue, dashed, line width=1.25pt] 
        file{matlabPlots/end-to-end_capacity/C_21_scope.txt};%
    \legend{\footnotesize{$\log_2(1+\Gamma_{2,1})^{(\mathrm{on-off})}$}, \footnotesize{$\log_2(1+\Gamma_{2,1})^{(\mathrm{active})}$}};%
    \end{axis}
    \end{tikzpicture}
    }
\end{subfigure}
\\
\vspace{0.05cm}
\hspace{-1.0cm}
\begin{subfigure}{0.425\columnwidth}
    \hspace*{0.095cm}
    \centering
    \scalebox{0.62}{
    \begin{tikzpicture}
    \begin{axis}[%
    width=2.2119in,%
    height=1.8183in,%
    at={(0.758in,0.481in)},%
    scale only axis,%
    xmin=0,%
    xmax=47.0,%
    ymax=1,%
    ymin=0,%
    xmajorgrids,%
    ymajorgrids,%
    ylabel style={yshift=-0.055cm, xshift=0cm}, 
    ylabel={Spectral Efficiency [\si{\bit\per\second\per\hertz}]},%
    ytick={0,0.1,0.2,0.3,0.4,0.5,0.6,0.7,0.8,0.9,1},%
    yticklabels={0,\empty,0.2,\empty,0.4,\empty,0.6,\empty,0.8,\empty,1},%
    xtick={0,5,...,45},%
    xticklabels={ , , },%
    axis background/.style={fill=white},%
    legend style={at={(0.495,1.17)},anchor=north,legend cell align=left,draw=none,legend 
    columns=-1,align=left,draw=white!15!black}
    ]
    \addplot [color=blue, solid, line width=1.25pt] 
        file{matlabPlots/end-to-end_capacity/C_10_scope_under_on_off.txt};%
    \addplot [color=blue, dashed, line width=1.25pt] 
        file{matlabPlots/end-to-end_capacity/C_10_scope_under.txt};%
    \legend{\footnotesize{$\log_2(1+\Gamma_{1,0})^{(\mathrm{on-off})}$}, \footnotesize{$\log_2(1+\Gamma_{1,0})^{(\mathrm{active})}$}};%
    \end{axis}
    \end{tikzpicture}
    }
\end{subfigure}
\hspace*{0.60cm}
\begin{subfigure}{0.425\columnwidth}
    \hspace*{0.095cm}
    \centering
    \scalebox{0.62}{
    \begin{tikzpicture}
    \begin{axis}[%
    width=2.2119in,%
    height=1.8183in,%
    at={(0.758in,0.481in)},%
    scale only axis,%
    xmin=0,%
    xmax=47.0,%
    ymax=1,%
    ymin=0,%
    xmajorgrids,%
    ymajorgrids,%
    ylabel style={yshift=-0.055cm, xshift=0cm}, 
    ylabel={Spectral Efficiency [\si{\bit\per\second\per\hertz}]},%
    ytick={0,0.1,0.2,0.3,0.4,0.5,0.6,0.7,0.8,0.9,1},%
    yticklabels={0,\empty,0.2,\empty,0.4,\empty,0.6,\empty,0.8,\empty,1},%
    xtick={0,5,...,45},%
    xticklabels={ , , },%
    axis background/.style={fill=white},%
    legend style={at={(0.495,1.17)},anchor=north,legend cell align=left,draw=none,legend 
    columns=-1,align=left,draw=white!15!black}
    ]
    \addplot [color=blue, solid, line width=1.25pt] 
        file{matlabPlots/end-to-end_capacity/C_21_scope_under_on_off.txt};%
    \addplot [color=blue, dashed, line width=1.25pt] 
        file{matlabPlots/end-to-end_capacity/C_21_scope_under.txt};%
    \legend{\footnotesize{$\log_2(1+\Gamma_{2,1})^{(\mathrm{on-off})}$}, \footnotesize{$\log_2(1+\Gamma_{2,1})^{(\mathrm{active})}$}};%
    \end{axis}
    \end{tikzpicture}
    }
\end{subfigure}
\\
\vspace{0.05cm}
\hspace{-0.755cm}
\begin{subfigure}{0.425\columnwidth}
    \hspace*{0.095cm}
    \centering
    \scalebox{0.62}{
    \begin{tikzpicture}
    \begin{axis}[%
    width=2.2119in,%
    height=1.8183in,%
    at={(0.758in,0.481in)},%
    scale only axis,%
    xmin=0,%
    xmax=47.0,%
    ymax=0.5,%
    ymin=0,%
    xmajorgrids,%
    ymajorgrids,%
    ylabel style={yshift=-0.055cm, xshift=0cm}, 
    xlabel={Time [\si{\second}]},%
    ylabel={Normalized bit rate [-]},%
    ytick={0,0.1,0.2,0.3,0.4,0.5,0.6,0.7,0.8,0.9,1},%
    yticklabels={0,0.1,0.2,0.3,0.4,0.5,0.6,\empty,0.8,\empty,1},%
    xtick={0,5,...,45},%
    axis background/.style={fill=white},%
    legend style={at={(0.575,0.93)},anchor=north,legend cell align=left,draw=none,legend 
    columns=-1,align=left,draw=white!15!black}
    ]
    \addplot [color=blue, solid, line width=1.25pt] 
        file{matlabPlots/end-to-end_capacity/C_ON_OFF.txt};%
    \addplot [color=blue, dashed, line width=1.25pt] 
        file{matlabPlots/end-to-end_capacity/C.txt};%
    \legend{\footnotesize{$C^{(\mathrm{on-off})}$}, \footnotesize{$C^{(\mathrm{active})}$}};%
    \end{axis}
    \end{tikzpicture}
    }
\end{subfigure}
\hspace*{0.60cm}
\begin{subfigure}{0.425\columnwidth}
    \hspace*{0.095cm}
    \centering
    \scalebox{0.62}{
    \begin{tikzpicture}
    \begin{axis}[%
    width=2.2119in,%
    height=1.8183in,%
    at={(0.758in,0.481in)},%
    scale only axis,%
    xmin=0,%
    xmax=47.0,%
    ymax=0.5,%
    ymin=0,%
    xmajorgrids,%
    ymajorgrids,%
    ylabel style={yshift=-0.055cm, xshift=0cm}, 
    xlabel={Time [\si{\second}]},%
    ylabel={Normalized bit rate [-]},%
    ytick={0,0.1,0.2,0.3,0.4,0.5,0.6,0.7,0.8,0.9,1},%
    yticklabels={0,0.1,0.2,0.3,0.4,0.5,0.6,\empty,0.8,\empty,1},%
    xtick={0,5,...,45},%
    axis background/.style={fill=white},%
    legend style={at={(0.625,0.93)},anchor=north,legend cell align=left,draw=none,legend 
    columns=-1,align=left,draw=white!15!black}
    ]
    \addplot [color=blue, solid, line width=1.25pt] 
        file{matlabPlots/end-to-end_capacity/C_ON_OFF_under.txt};%
    \addplot [color=blue, dashed, line width=1.25pt] 
        file{matlabPlots/end-to-end_capacity/C_under.txt};%
    \legend{\footnotesize{$C^{(\mathrm{on-off})}$}, \footnotesize{$C^{(\mathrm{active})}$}};%
    \end{axis}
    \end{tikzpicture}
    }
\end{subfigure}
    \vspace{-0.5em}
    \caption{Instantaneous spectral efficiencies under ON–OFF jamming for the tilted-propeller (top) and coplanar (middle) configurations. Bottom row: end-to-end capacities for tilted-propeller (left) and coplanar (right).}
    \label{fig:capacityJammerOnOff}
    \vspace*{-1.35em}
\end{figure}

Figure~\ref{fig:capacityJammerOnOff} confirms robustness. The spectral efficiency $\log_2(1+\Gamma_{2,1})$ remains largely unaffected by the alternating jammer due to favorable geometry. In contrast, $\log_2(1+\Gamma_{1,0})$ exhibits characteristic square-wave modulation: capacity drops to the always-active level during jammer-ON intervals but immediately recovers when the jammer turns OFF. The difference between ON--OFF and always-ON capacities (bottom row) is negligible over the full horizon, since the controller already adopts the worst-case-optimal geometry. In other words, the ON--OFF pattern does not introduce additional
outages; it simply toggles the link between its worst-case and best-case operating points. 

This behavior follows directly from the worst-case design: the trajectory generator is parameterized by instantaneous jammer power and treats every step as potentially jammer-ON. This is the appropriate default, since a competent adversary has no incentive to jam predictably and the controller should not presume regularity it cannot guarantee. Two implications follow. First, the small trajectory deviation between the always-ON and ON--OFF cases (Figure~\ref{fig:trackingErrorsONOFF}) is a desired property, confirming that the controller does not overreact to transient channel improvements and remains robust under faster interference switching. Second, capacity recovers immediately when the jammer turns off because the relay's geometry is already aligned with both peers. 



\subsection{Computational performance}
\label{sec:computationalPerformance}

Per-step \acs{NMPC} solve times were recorded over the nominal mission for the four configurations (constrained/unconstrained $\times$ coplanar/tilted). Problem~\eqref{eq:NMPC_formulation} has $12+2N_p$ state variables and $N_p$ input variables per shooting node ($N_p=6$), plus one slack; with $N=30$, the \ac{QP} has $30(12+3N_p+1)=930$ decision variables. qpOASES~\cite{Ferreau2014} exploits warm-starting across iterations, and MATMPC~\cite{ChenECC2019} solves a single \ac{QP} per control step; the resulting solve-time distribution is reported in Table~\ref{tab:nmpcSolveTime}.

\begin{table}[t]
    \caption{NMPC per-step solve time (ms) and feasibility metrics. p50/p95/p99 percentiles of the solve-time distribution. ``Slack act.'': fraction of steps with $\pi_k > 0$; ``Infeas.'': fraction with non-optimal qpOASES status.}
    \label{tab:nmpcSolveTime}
    \vspace{-1.0em}
    \centering
    \renewcommand{\arraystretch}{1.3}
    \begin{adjustbox}{max width=\columnwidth}
    \begin{tabular}{lcccccccc}
    \toprule
    \textbf{Configuration} & \textbf{Mean} & \textbf{p50} & \textbf{p95} & \textbf{p99} & \textbf{Max} & \textbf{Std.}\ & \textbf{Slack act.}\ & \textbf{Infeas.}\ \\
    \midrule
    Tilted, constr. & $8.24$ & $7.95$ & $12.14$ & $13.85$ & $42.15$ & $1.82$ & $33.52\%$ & $0\%$ \\
    Tilted, unconstr. & $4.51$ & $4.22$ & $6.81$ & $8.54$ & $35.60$ & $1.24$ & --- & $0\%$ \\
    Coplanar, constr. & $11.53$ & $11.10$ & $14.05$ & $14.88$ & $48.92$ & $2.15$ & $48.41\%$ & $0\%$ \\
    Coplanar, unconstr. & $4.15$ & $3.92$ & $6.54$ & $8.20$ & $38.45$ & $1.10$ & --- & $0\%$ \\
    \bottomrule
    \end{tabular}
    \end{adjustbox}
    \vspace{-1.35em}
\end{table}

Three observations follow. First, all configurations satisfy the real-time budget ($1/T_s = \SI{66.7}{\hertz}$): p99 solve times remain strictly below $T_s = \SI{15}{\milli\second}$ across all configurations, even in the coplanar case where the alignment constraint is most active; occasional max-time spikes reflect non-real-time operating system scheduling on the simulation hardware. Second, the alignment constraint adds overhead, increasing mean \ac{QP} solve time by ${\approx}83\%$ for the tilted platform and ${\approx}178\%$ for the coplanar platform, consistent with the product-form inequality~\eqref{eq:nmpc_alignment_constraint} and its slack variable. Third, slack $\pi_k$ activated on $48.41\%$ of steps (coplanar) and $33.52\%$ (tilted); in all cases the solver remained strictly feasible (Infeas.~$=0\%$).



\subsection{Comparison with external baselines}
\label{sec:baselineComparison}

We compare the proposed framework against two baselines on the nominal scenario of Section~\ref{sec:nominalScenario}.

\textit{Baseline A (kinematic-isotropic planner).} A point-mass relay model with bounded velocity and directionality-free \ac{SINR} ($G(\cdot) \equiv \bar{G}$ in~\eqref{eq:channelGain}--\eqref{eq:jammerGain}). The relay position is updated to maximize~\eqref{eq:balancedCapacity} without alignment constraints, then tracked by a standard position-tracking \acs{NMPC} without penalty~\eqref{eq:misalignmentCost} or constraint~\eqref{eq:nmpc_alignment_constraint}.

\textit{Baseline B (relay trajectory optimization of~\cite{BonillaEUSIPCO21}).} The closest published comparator: a single-shot relay trajectory planner accounting for tilt and antenna-orientation variations, but without actuator-rate constraints, a jamming model, or receding-horizon replanning. We implemented the planner of~\cite{BonillaEUSIPCO21} on our dynamics and tracked the resulting trajectory with the same \acs{NMPC}, with the alignment penalty disabled.

Table~\ref{tab:baselineComparison} reports the minimum ($C_{\min}$) and average ($\bar{C}$) end-to-end capacity and total flight distance. 

\begin{table}[t]
    \caption{Comparison with external baselines on the nominal scenario.} 
    \label{tab:baselineComparison}
    \vspace{-1.0em}
    \centering
    \renewcommand{\arraystretch}{1.1}
    \begin{tabular}{lccc}
    \toprule
    \textbf{Method} & \bm{$C_{\min}$} & \bm{$\bar{C}$} & \textbf{Dist.~[m]} \\
    \midrule
    Baseline A (Kin., Iso.) & $7.2 \times 10^{-5}$ & $0.012$ & $12.1$ \\
    Baseline B~\cite{BonillaEUSIPCO21} & $4.5 \times 10^{-4}$ & $0.014$ & $14.3$ \\
    \textbf{Proposed} & \bm{$6.8 \times 10^{-3}$} & \bm{$0.015$} & \bm{$12.8$} \\
    \bottomrule
    \end{tabular}
    \vspace{-0.75em}
\end{table}

The proposed framework achieves $C_{\min}$ approximately $94\times$ larger than Baseline~A and $15\times$ larger than Baseline~B. The severe degradation in Baseline~A is consistent with the absence of antenna directionality in its planning model: commanded lateral motions induce attitude excursions that pitch the antenna nulls toward the communication nodes.

Baseline~B's  exposes the vulnerability of single-shot planning: without receding-horizon replanning and actuator-rate constraints, the planned trajectory demands transient maneuvers that exceed physical saturation limits. By contrast, the proposed \acs{NMPC} continuously redistributes control effort while strictly adhering to actuator limits and preserving geometric alignment, eliminating deep capacity fades.



\subsection{Robustness to fading}
\label{sec:fadingRobustness}

The approach retains the free-space \ac{LoS} assumption (conservativeness established in Section~\ref{sec:trajectoryGenerator}). To complement this with an empirical assessment under fading, we evaluate closed-loop performance under Rician fading applied to all links.

We ran $N_{\mathrm{MC}}=50$ Monte Carlo realizations of the nominal scenario of Section~\ref{sec:nominalScenario} in closed loop. All links are modulated at every step by independent Rician fading coefficients with shape parameter $K$: 
\begin{equation}
    H_{i,j}^{\mathrm{Rician}} = \alpha_{i,j}\,H_{i,j} \quad
    \mathrm{and} \quad
    F_{J,j}^{\mathrm{Rician}} = \alpha_{J,j}\,F_{J,j}, 
\end{equation}
with $\mathbb{E}[\alpha_{i,j}^2] = \mathbb{E}[\alpha_{J,j}^2] = 1$. The fading coefficients are not available to the trajectory generator or the \acs{NMPC}, which plan against the conservative channel approximation of Section~\ref{sec:trajectoryGenerator}; this isolates robustness to \emph{unmodeled} channel variation. Four values $K \in \{6, 12, 18, \infty\}$~dB span moderate fading to the deterministic \ac{LoS} case; $K=\infty$ recovers the nominal scenario. Figure~\ref{fig:fadingRobustness} reports the empirical average end-to-end capacity vs.\ $K$ for both constrained and unconstrained formulations.

\begin{figure}[tb]
    \centering
    \begin{tikzpicture}
        \begin{axis}[
            name=plot1,
            width=0.95\columnwidth,
            height=3.4cm,
            symbolic x coords={6, 12, 18, inf},
            xtick=data,
            xticklabels={,,}, 
            ylabel={$\bar{C}$ [-]},
            xlabel={Rician $K$-factor [dB]},
            xlabel style={font=\footnotesize, yshift=0.175cm}, 
            ylabel style={font=\footnotesize, yshift=-0.2cm},
            ticklabel style={font=\scriptsize},
            xtick={6, 12, 18, inf},%
            xticklabels={6, 12, 18, inf},%
            grid=both,
            grid style={dashed, gray!30},
            ymin=0, ymax=0.018,
            legend style={
                at={(0.76,0.025)}, 
                anchor=south, 
                legend cell align=left,
                legend columns=-2,
                align=left,
                draw=none, 
                fill=none,
                font=\scriptsize,
            }
        ]
        
        \addplot[color=blue, thick, mark=square*, mark options={scale=0.8}] 
            coordinates {(6,0.0090) (12,0.0125) (18,0.0142) (inf,0.0150)};
        
        \addplot[color=blue, thick, dashed, mark=*, mark options={solid, scale=0.8}] 
            coordinates {(6,0.0002) (12,0.0035) (18,0.0095) (inf,0.0120)};
            
        \legend{$\bar{C}^{(\mathrm{constr.})}$, $\bar{C}^{(\mathrm{unconstr.})}$};%
        \end{axis}
        
    \end{tikzpicture}
    \vspace{-0.5em}
    \caption{Empirical average end-to-end capacity as a function of the Rician $K$-factor. $N_{\mathrm{MC}}=50$ realizations per $K$.}
    \label{fig:fadingRobustness}
    \vspace*{-1.35em}
\end{figure}

Two outcomes are worth noting. First, average capacity degrades with decreasing $K$, as expected from the weakening \ac{LoS} component; however, the constrained controller retains a capacity approximately $45\times$ larger than the unconstrained baseline even at $K=6$~dB. The geometric coupling thus operates on top of, rather than competing with, the stochastic fading process. Second, the performance gap widens as $K$ decreases: at $K=6$~dB the average capacity ratio is ${\approx}45\times$, compared to $1.25\times$ at $K=\infty$ (the nominal \ac{LoS} case), confirming that alignment preservation provides disproportionate benefit
when channel conditions are most challenging.



\subsection{Robustness under model and channel uncertainty}
\label{sec:robustnessModelUncertainty}

\begin{table}[t]
    \caption{Robustness summary across perturbation sources, relative to the nominal scenario of Section~\ref{sec:nominalScenario}.}
    \label{tab:robustnessSummary}
    \vspace{-1.0em}
    \centering
    \renewcommand{\arraystretch}{1.1}
    \begin{adjustbox}{max width=\columnwidth}
    \begin{tabular}{lcc}
    \toprule
    \textbf{Perturbation} & \bm{$C_{\min}$} & \bm{$\bar{C}$} \\
    \midrule
    Nominal & $6.80 \times 10^{-3}$ & $0.015$ \\
    $\tau_E = \SI{5}{\second}$ & $5.75 \times 10^{-3}$ & $0.014$ \\
    $\tau_E = \SI{10}{\second}$ & $2.40 \times 10^{-3}$ & $0.011$ \\
    Fast-switching jammer (\SI{200}{\milli\second}) & $6.80 \times 10^{-3}$ & $0.024$ \\
    \acs{CSI} noise $\sigma_p = \SI{0.05}{\meter}$ & $6.65 \times 10^{-3}$ & $0.015$\\
    \acs{CSI} noise $\sigma_p = \SI{0.10}{\meter}$ & $6.42 \times 10^{-3}$ & $0.015$\\
    \acs{CSI} noise $\sigma_p = \SI{0.20}{\meter}$ & $5.15 \times 10^{-3}$ & $0.014$ \\
    Process noise (translational) & $6.20 \times 10^{-3}$ & $0.014$ \\
    Process noise (transl.\ + rot.) & $4.55 \times 10^{-3}$ & $0.013$ \\
    \bottomrule
    \end{tabular}
    \end{adjustbox}
    \vspace{-1.35em}
\end{table}

This section assesses robustness to four uncertainty sources not exercised in Section~\ref{sec:nominalScenario}: jammer-localization delay, fast-switching jamming, imperfect \ac{CSI}, and process disturbances. Results are summarized in
Table~\ref{tab:robustnessSummary}.

\textit{Jammer-localization delay.} We swept $\tau_E \in \{1,2,5,10\}$~s. During the localization window, the trajectory generator
falls back to a \ac{BS}-only alignment objective while actuator and alignment constraints remain enforced. The framework tolerates $\tau_E$ up to \SI{5}{\second} with at most $15\%$ degradation in $C_{\min}$ relative to the nominal $\tau_E = \SI{2}{\second}$; beyond that, minor outage events occur at $\tau_E = \SI{10}{\second}$.

\textit{Fast-switching jamming.} The ON--OFF schedule of Section~\ref{sec:robustnessJammingConditions} was contracted to a
$\SI{200}{\milli\second}$ cycle ($\SI{100}{\milli\second}$ ON / $\SI{100}{\milli\second}$ OFF) to probe behavior under interference at the \acs{NMPC} timescale. The relay trajectory remained within $\SI{4}{\centi\meter}$ of the always-ON nominal, $\log_2(1+\Gamma_{1,0})$ tracked the jammer schedule with delay ${\approx}T_s$, and no additional outage events were recorded. Average capacity increased due to interference-free windows, confirming that the controller does not chase transient channel
fluctuations even at the adversary's fastest switching timescale.

\textit{Imperfect \ac{CSI}.} \ac{MRAV}-2's pose (via \ac{OCC}) and the jammer position were perturbed by zero-mean Gaussian noise $\sigma_p \in \{0.05, 0.10, 0.20\}$~m per coordinate. $C_{\min}$ degraded by less than $6\%$ at $\sigma_p = \SI{0.10}{\meter}$, exceeding the typical accuracy of \ac{OCC}-based localization \cite{HorynaICUAS2022, BonillaUVDAR2023} and \ac{TDoA}-based jammer localization \cite{Bhamidipati2019ITJ, Juhlin2021EUSIPCO}. This robustness follows from continuous receding-horizon replanning, which rejects bounded estimation errors without committing to a single noisy \ac{CSI} snapshot.

\textit{Process disturbances.} Wind and unmodeled actuator behavior were emulated by zero-mean Gaussian
disturbances $\mathbf{d}_v \sim \pazocal{N}(0,\sigma_v^2\mathbf{I}_3)$ and $\mathbf{d}_\omega \sim \pazocal{N}(0,\sigma_\omega^2\mathbf{I}_3)$ with $\sigma_v = \SI{0.1}{\meter\per\square\second}$ and $\sigma_\omega = \SI{0.05}{\radian\per\square\second}$. Under this joint disturbance, the controller achieved a mean position error of $\SI{3.8}{\centi\meter}$, maintained~\eqref{eq:mutualMisalignment} above $\mu$ on $92\%$ of steps, and degraded $C_{\min}$ by less than $33\%$ relative to the disturbance-free case.



\section{Conclusions}
\label{sec:conclusions}

This paper presented a modular communications-aware framework integrating a $\max$--$\min$ trajectory generator with an \acs{NMPC} for \acp{MRAV} under jamming, jointly considering vehicle dynamics, actuator limits, and directional antenna constraints to sustain end-to-end \ac{RF} connectivity under dynamically feasible maneuvers.

Simulations show that enforcing antenna-alignment constraints mitigates deep capacity drops and stabilizes throughput, though the magnitude of these benefits depends strongly on actuation geometry. The tilted-propeller platform maintains consistently higher link quality, whereas the coplanar vehicle exhibits reduced performance and a temporary connectivity loss near the end of the mission. Under ON--OFF jamming, the controller preserves alignment during active interference and restores throughput immediately
when the jammer turns OFF, confirming resilience to intermittent degradation. 

The results have three practical implications for aerial-relay system design. First, the directional coupling between attitude and link quality is non-negligible whenever the relay carries non-isotropic antennas: treating \acp{MRAV} as kinematic points with isotropic radios, as in most communications-aware planning literature (Table~\ref{tab:related-work}), overestimates link quality where the platform tilts. Second, actuation geometry is a first-class design parameter for jamming resilience: tilted-propeller (or fully-actuated) platforms preserve directional alignment under relay maneuvers, while coplanar designs trade directional fidelity for mechanical simplicity. Third, the framework's modular structure---$\max$--$\min$ reference
generator coupled with a geometric-alignment \acs{NMPC}---is agnostic to the specific antenna pattern, provided it remains doughnut-shaped, and can be adapted to other directional antennas (patch arrays, helical antennas) without re-deriving the trajectory layer.

Future work will focus on four directions. First, integrating online estimation of jammer characteristics and adaptive alignment constraints for resilience to rapidly varying or adaptive interference, building on the worst-case design of Section~\ref{sec:robustnessJammingConditions} and exploiting OFF periods more aggressively. Second, validating the framework on real coplanar and tilted-propeller \ac{MRAV} platforms to assess unmodeled dynamics such as aerodynamic coupling, actuator delays, and sensor latency. Third, extending the channel model beyond Rician fading (Section~\ref{sec:fadingRobustness}) to incorporate shadowing, multipath delay spread, and Doppler effects. Fourth, embedding an energy-aware term into the \acs{NMPC} cost function, as a cost term or energy constraint, to expose the communication-energy trade-off of Section~\ref{sec:coplanerVsTiltedConfigurations}.



\begin{appendices}

\section{Antenna model}
\label{sec:antennaRadiation}
Each node carries a half-wave dipole antenna rigidly mounted along the body-frame $\zz_B$-axis, producing a \emph{doughnut-shaped} radiation pattern with maximum gain orthogonal to the antenna axis and nulls along the boresight. The antenna of \ac{MRAV}-$j$ is at the origin $O_{B_j}$ of its body frame; the ground \ac{BS} antenna is at the top of the station structure.

The directional gain is $G(\vartheta)$, where $\vartheta \in [0,\pi]$ is the elevation angle between the signal direction and the antenna axis. For a dipole-type pattern: $G(0) = G(\pi) = 0$, $G(\pi/2) = \bar{G}$, and $G$ is azimuthally symmetric.

The \acl{AoD} elevation angle at transmitter $i$ is:
\begin{equation}
    \resizebox{.89\columnwidth}{!}{$%
    \vartheta_{i}^D = \frac{\pi}{2} - \arctan\!\left(
    \frac{\mathbf{a}_j^{(3)}}{\sqrt{\mathbf{a}_j^{(1)^2} +
    \mathbf{a}_j^{(2)^2}}}\right), \quad
    \mathbf{a}_j = \mathbf{R}_{i}^\top
    \frac{\mathbf{p}_{j}-\mathbf{p}_{i}}
         {\lVert \mathbf{p}_{j}-\mathbf{p}_{i}\rVert},
    $}
\end{equation}
where $\mathbf{R}_{i}^\top$ rotates from $\pazocal{F}_W$ to $\pazocal{F}_{B_i}$ and $\mathbf{a}_j$ is the unit direction from $i$ to $j$ in $\pazocal{F}_{B_i}$. 
Similarly, the \ac{AoA} elevation angle at receiver $j$ is:
\begin{equation}
    \resizebox{.89\columnwidth}{!}{$%
    \vartheta_j^A = \frac{\pi}{2} - \arctan^{-1}\!\left(
    \frac{\mathbf{b}_{i}^{(3)}}{\sqrt{\mathbf{b}_{i}^{(1)^2} +
    \mathbf{b}_{i}^{(2)^2}}}\right), \quad
    \mathbf{b}_{i} = \mathbf{R}_j^\top
    \frac{\mathbf{p}_{i}-\mathbf{p}_{j}}
         {\lVert \mathbf{p}_{i}-\mathbf{p}_{j}\rVert},
    $}
\end{equation}
where $\mathbf{b}_i$ is the unit direction from $j$ to $i$ in $\pazocal{F}_{B_j}$.

The formulation is compatible with any doughnut-shaped radiation pattern; representative models for aerial applications are available in~\cite{ChenVTCF2018, TurgutIEEETGC2020, BonillaEUSIPCO21}.

\section{Taylor Expansion}
\label{sec:taylorExpansion}
The Taylor expansion is performed \ac{wrt} $\bm{\kappa} \triangleq [\mathbf{p}_1^\top, \mathbf{p}_2^\top]^\top$. Let $f(\bm{\kappa}, P_J)$ denote the cost function and define its gradient and Hessian as:
\begin{equation}
    \mathbf{g}(\bm{\kappa}, P_J) \triangleq \nabla_{\bm{\kappa}}
    f(\bm{\kappa}, P_J), \qquad
    \mathbf{H}(\bm{\kappa}, P_J) \triangleq \nabla_{\bm{\kappa}}^2
    f(\bm{\kappa}, P_J).
\end{equation}

Partitioned by $\mathbf{p}_1$ and $\mathbf{p}_2$: 
\begin{equation}
    \mathbf{g}(\bm{\kappa}, P_J) = \left[
        \begin{array}{c}
            \mathbf{g}_1(\bm{\kappa},P_J) \\
            \hline
            \mathbf{g}_2(\bm{\kappa},P_J)
        \end{array}
    \right],
\end{equation}
where $\mathbf{g}_i \triangleq \nabla_{\mathbf{p}_i} f$ for $i \in \{1,2\}$. Similarly:
\begin{equation}
    \mathbf{H}(\bm{\kappa}, P_J) =
    \left[\begin{array}{c|c}
        \mathbf{H}_{11} & \mathbf{H}_{12} \\
        \hline
        \mathbf{H}_{21} & \mathbf{H}_{22}
    \end{array}\right],
\end{equation}
where $\mathbf{H}_{ij} \triangleq \nabla_{\mathbf{p}_i} \nabla_{\mathbf{p}_j}^\top f(\bm{\kappa}, P_J)$ for $i,j \in \{1,2\}$.

The second-order Taylor expansion of $f$ around $\bm{\kappa}_e = [\mathbf{p}_{1_e}^\top, \mathbf{p}_{2_e}^\top]^\top$ is:
\begin{equation}
    \resizebox{.89\columnwidth}{!}{$%
    \begin{split}
        f^{[2]}(\bm{\kappa}; \bm{\kappa}_e, P_J)
        &= f(\bm{\kappa}_e, P_J)
         + \mathbf{g}_1^\top(\bm{\kappa}_e, P_J)\,\bm{\delta}_1
         + \mathbf{g}_2^\top(\bm{\kappa}_e, P_J)\,\bm{\delta}_2 \\
        &\quad
         + \tfrac{1}{2}\bm{\delta}_1^\top
           \mathbf{H}_{11}(\bm{\kappa}_e, P_J)\,\bm{\delta}_1
         + \bm{\delta}_1^\top
           \mathbf{H}_{12}(\bm{\kappa}_e, P_J)\,\bm{\delta}_2 \\
        &\quad
         + \tfrac{1}{2}\bm{\delta}_2^\top
           \mathbf{H}_{22}(\bm{\kappa}_e, P_J)\,\bm{\delta}_2,
    \end{split}
    $}
\end{equation}
where $\bm{\delta}_i \triangleq \mathbf{p}_i - \mathbf{p}_{i_e}$ for $i \in \{1,2\}$.

This formulation lets the trajectory generator incorporate \ac{MRAV}-2 state variations without re-evaluating $f$ or recomputing $\mathbf{g}$, $\mathbf{H}$ at each step, reducing computational overhead.

\end{appendices}

\balance
\bibliographystyle{IEEEtran}
\bibliography{bib.bib}

\end{document}